\documentclass[conference]{IEEEtran}
\usepackage{times}
\usepackage{cite}
\usepackage{multicol}
\usepackage{graphicx}
\usepackage{amsfonts}
\usepackage{mathptmx} 
\usepackage{times} 
\usepackage{amsmath} 
\usepackage{amssymb}  
\usepackage{array}
\usepackage{colortbl}
\usepackage[T1]{fontenc}
\usepackage{wrapfig,lipsum}
\usepackage{booktabs}
\usepackage{bigstrut}
\usepackage{multirow}
\usepackage{placeins}
\usepackage{gensymb}
\usepackage{xcolor}
\usepackage{soul}
 
\DeclareMathOperator*{\argmaxB}{argmax} 
\usepackage{tikz}
\usepackage{yfonts}
\usepackage{graphicx}
\usepackage{import}
\usepackage{newtxtext}
\usepackage{newtxmath}
\usepackage{hhline}
\usepackage{stfloats}
\usepackage{setspace}
\usepackage{bigstrut}
\usepackage{atbegshi,picture}   
\usepackage{mathtools}
\usepackage[title]{appendix}
\usepackage[linesnumbered,ruled]{algorithm2e}

\usepackage[bookmarks=true,linkcolor=blue,colorlinks = true]{hyperref}

\DeclareMathAlphabet{\mathcal}{OMS}{cmsy}{m}{n}
\newcommand{\mc}[1]{\mathcal{#1}}

\definecolor{comment_green}{rgb}{0.55,0.78,0.243}

\SetCommentSty{mycommfont}

\definecolor{red}{RGB}{255,0,0}

\newenvironment{myitem}{\begin{list}{$\bullet$}
{\setlength{\itemsep}{-0pt}
\setlength{\topsep}{0pt}
\setlength{\labelwidth}{0pt}
\setlength{\leftmargin}{10pt}
\setlength{\parsep}{-0pt}
\setlength{\itemsep}{0pt}
\setlength{\partopsep}{0pt}}}%
{\end{list}}

\hypersetup{pdfauthor={Bowen Wen, Wenzhao Lian, Kostas Bekris and Stefan Schaal},
            pdftitle={You Only Demonstrate Once: Category-Level Manipulation from Single Visual Demonstration},
            pdfsubject={robotics; computer vision; artificial intelligence},
            pdfkeywords={Manipulation;Imitation Learning;Perception},
}

\begin{document}

\title{You Only Demonstrate Once: Category-Level Manipulation from Single Visual Demonstration}



%
\author{\authorblockN{Bowen Wen\authorrefmark{4}\authorrefmark{2},
Wenzhao Lian\authorrefmark{2},
Kostas Bekris\authorrefmark{4} and 
Stefan Schaal\authorrefmark{2}}

\authorblockA{\authorrefmark{2}Intrinsic Innovation LLC, CA, USA.\\
{Email: \{wenzhaol, sschaal\}@intrinsic.ai}}
\authorblockA{\authorrefmark{4}Department of Computer Science, Rutgers University, NJ, USA \\ 
Email: \{bw344, kostas.bekris\}@cs.rutgers.edu}
}

\AtBeginShipoutNext{\AtBeginShipoutUpperLeft{%
  \put(\dimexpr\paperwidth-0.4cm\relax,-0.6cm){\makebox[0pt][r]{\framebox{\scriptsize Accepted in Robotics: Science and Systems (RSS) 2022}}}%
}}


\twocolumn[{
\renewcommand\twocolumn[1][]{#1}%
\maketitle
\begin{center} 
    \centering
     \includegraphics[width = 0.91\textwidth]{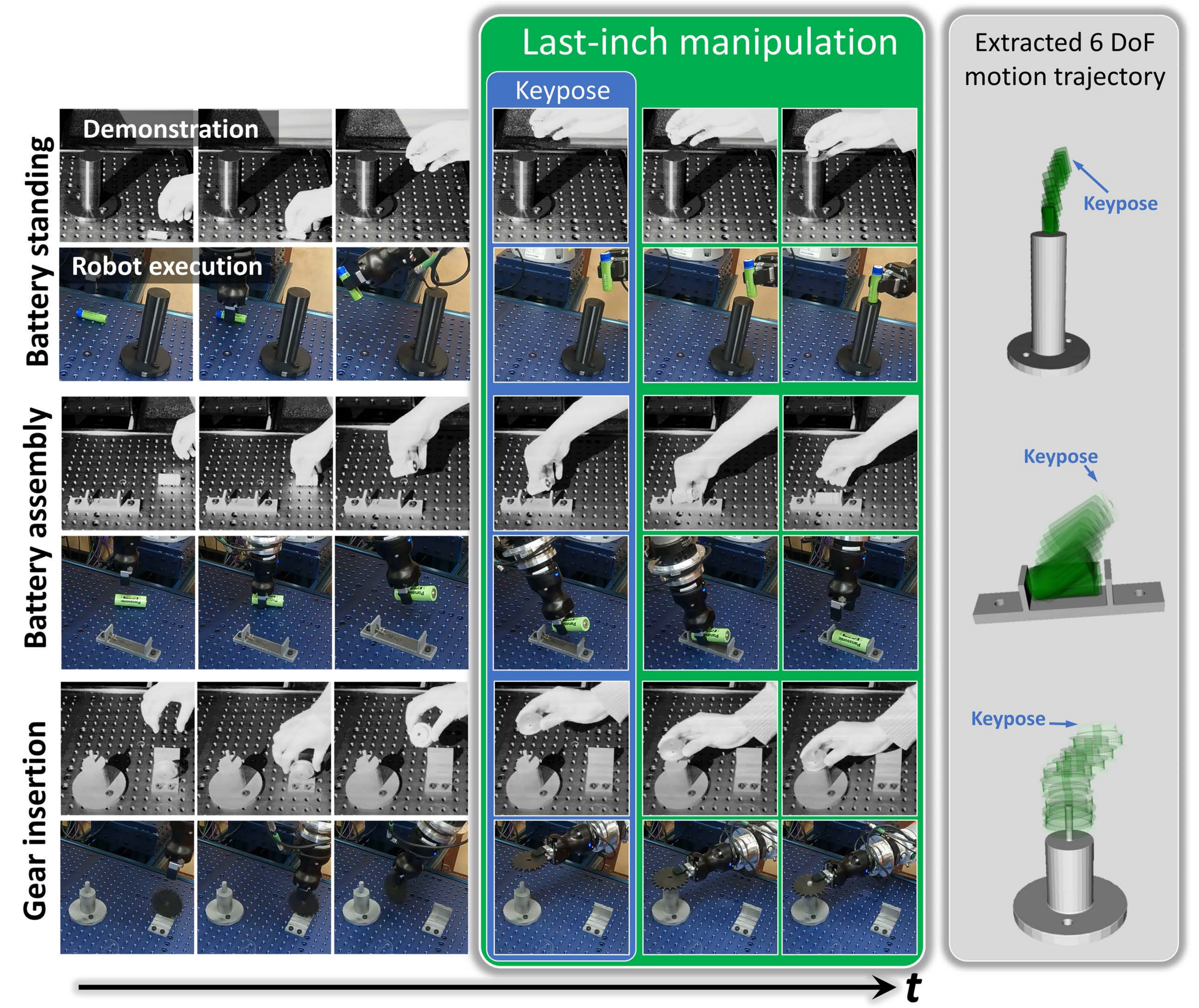}
    \vspace{-.05in}
\end{center}%
}]

\begin{abstract}
Promising results have been achieved recently in category-level manipulation that generalizes across object instances. Nevertheless, it often requires expensive real-world data collection and manual specification of semantic keypoints for each object category and task. Additionally, coarse keypoint predictions and ignoring intermediate action sequences hinder adoption in complex manipulation tasks beyond pick-and-place. This work proposes a novel, category-level manipulation framework that leverages an object-centric, category-level representation and model-free 6 DoF motion tracking. The canonical object representation is learned solely in simulation and then used to parse a category-level, task trajectory from a single demonstration video. The demonstration is reprojected to a target trajectory tailored to a novel object via the canonical representation. During execution, the manipulation horizon is decomposed into long-range, collision-free motion and last-inch manipulation. For the latter part, a category-level behavior cloning (CatBC) method leverages motion tracking to perform closed-loop control.  CatBC follows the target trajectory, projected from the demonstration and anchored to a dynamically selected category-level coordinate frame. The frame is automatically selected along the manipulation horizon by a local attention mechanism. This framework allows to teach different manipulation strategies by solely providing a single demonstration, without complicated manual programming. Extensive experiments demonstrate its efficacy in a range of challenging industrial tasks in high-precision assembly, which involve learning complex, long-horizon policies. The process exhibits robustness against uncertainty due to dynamics as well as generalization across object instances and scene configurations. The supplementary video is available at \url{https://www.youtube.com/watch?v=WAr8ZY3mYyw}
\end{abstract} 


\IEEEpeerreviewmaketitle

\section{INTRODUCTION}

 
Significant progress has been achieved in robotic manipulation for known objects \cite{lian2021benchmarking,kyrarini2019robot,andrewbwrss}, such as methods for acquiring and encoding task-relevant object knowledge. These methods range from training 6D pose estimators with CAD models to end-to-end reinforcement learning from repeated robot interaction with the task object. While manipulation skills for the exact object instance can be acquired with either strategy, it requires significant time and effort to transfer these skills to similar but novel instances. Additionally, the exact object instance is often unavailable until task execution, particularly in less structured environments.


This has motivated recent, promising results in improving the generalizability of robotic manipulation by learning category-level representations, such as semantic keypoints or dense correspondence \cite{florence2018dense,manuelli2019kpam}, where transferring manipulation skills across instances is formulated as aligning the semantic keypoints between intra-class object instances \cite{manuelli2019kpam}.
This direction has a few limitations, however: 

\noindent \textbf{Manipulation Task Complexity}: Trajectory optimization with manually specified goals and constraints has been demonstrated on simple tasks, such as pick-and-place and board-wiping \cite{manuelli2019kpam,gao2021kpam}. This process, however, becomes non-intuitive and tedious for more complex and long-horizon tasks. For instance, placing the battery into a spring-loaded charge device requires a sequence of actions, such as pressing towards one end along an angle, then aligning the battery with both ends and pressing down. In such cases, segmenting the manipulation action sequence, and manually specifying the goal and constraints for each segment are challenging. 



\noindent \textbf{Robustness}: Often in existing work, the manipulated object is assumed to remain static relative to the gripper at the time of the grasp and during manipulation. This assumption enables open-loop execution given only the initially detected keypoints \cite{manuelli2019kpam} and forward kinematics (FK) \cite{manuelli2019kpam,gao2021kpam}. Although force sensing is incorporated in prior work \cite{gao2021kpam}, it is not always available and typically handles limited local disturbances. When visual sensing is the primary modality, as in the setting of the current work, open-loop manipulation becomes less reliable, especially in long-horizon manipulation scenarios, as will be shown in the experiments (Sec. \ref{sec:exp}).



\noindent \textbf{Time and cost}: To ensure that the training distribution covers the category for learning correspondences, a large number of diverse object instances need to be collected and manually configured for scanning. Multi-view data collection, even when performed by a robot, is time-consuming and cumbersome. The same is true for human annotation of semantic keypoints.

To address these limitations, this work proposes closed-loop, category-level manipulation framework based exclusively on visual feedback, which can be applied to novel objects fast and inexpensively. Leveraging state-of-the-art solutions for model-free 6 DoF object motion tracking, manipulation trajectories are automatically extracted from a single demonstration video. The extracted demonstration trajectory is then represented in a category-level canonical space, which learns solely over synthetic data and enables transferring the manipulation skill across intra-class objects. The manipulation skill is also readily transferable across different task configurations by relying on self-adaptive local correspondences, which are regularized via an attention mechanism. During online robot execution, model-free 6 DoF object tracking is again used for visual feedback to aid the category-level behavior cloning process that guides the robot to follow the canonical trajectory for the target object. Overall, the contributions of this work can be summarized as follows:
\begin{myitem}
\item A novel, category-level manipulation framework leveraging object-centric representations trained solely in simulation and model-free 6 DoF object tracking. It achieves robustness and high-precision using only visual feedback. The manipulation skills are transferred across category instances via a Category-level Behavior Cloning (CatBC) process.
\item The framework is enabled by one-shot imitation learning where only a single third-person-view video demonstration is used. By virtue of the framework's modular design, the acquired skill is also generalizable to different environments and task configurations. Additionally, it allows quickly teaching the robot with different manipulation strategies without otherwise complicated manual programming.
\item An attention mechanism is proposed for dynamic category-level coordinate frame selection. It automatically and dynamically identifies the task-relevant local part of the object for the manipulation task and anchors the category-level, canonical frame for more fine-grained cross-instance alignment when performing behavior cloning. 
\item This work focuses on challenging manipulation tasks that require high precision and long-horizon actions. Extensive real-world experiments demonstrate significantly superior performance of the proposed framework compared to alternatives for category-level manipulation, in reliability, robustness, and training cost.

\end{myitem}

\section{RELATED WORK}

\noindent\textbf{Category-Level Manipulation} aims to learn manipulation skills that generalize across instances in the same category. During testing, it is expected to be readily applicable to novel instances, without the need for CAD models or additional robot-object interactions. To achieve this, representative work learns correspondences shared among similar object instances via dense pixel-wise representation \cite{florence2018dense,yang2021learning,chai2019multi} or semantic keypoints \cite{manuelli2019kpam,qin2020keto}. In particular, sparse semantic keypoint representations are often assigned task-relevant semantic priors via human annotation \cite{manuelli2019kpam,qin2020keto,vecerik2020s3k}. It is cumbersome, however, to manually specify semantic keypoints for each task and object category.  To circumvent this annotation effort, a dense correspondence model was recently proposed, together with self-supervised training over 2D image pairs acquired with a camera-mounted robot \cite{florence2018dense}.
This work aims to avoid manual or time-consuming processes. Instead of reasoning on 2D image pairs that are constrained to specific views, the proposed category-level, object-centric representation allows to directly reason in 3D space. It also imposes an explicit mapping among object instances as training supervision. Therefore, more reliable dense correspondence can be established so as to achieve higher precision than those based on contrastive learning \cite{florence2018dense, yang2021learning, chai2019multi}, as shown in the experiments (Sec. \ref{sec:exp}). 

\noindent\textbf{Behavior Cloning (BC)} collects expert demonstrations, and learns a policy taking as input observations and output actions \cite{osa2018algorithmic}. BC methods can be categorized into model-based \cite{grimes2006dynamic,englert2013probabilistic,nair2017combining} and model-free methods. The latter, which do not estimate dynamic models, can be further grouped into two types: policy learning \cite{duan2017one,stadie2017third,finn2017one} and trajectory learning \cite{schaal2005learning,calinon2009statistical,schulman2016learning}. More related to this work are vision-based BC methods that aim to learn manipulation skills from demonstration videos \cite{florence2019self,paradis2021intermittent,johns2021coarse,wu2020squirl,huang2019neural,liang2022learning}. In addition to video streams, they often rely on robot action labels acquired via extensive robot-object interaction. In contrast, the proposed approach is object-centric and eliminates this requirement. Related work aims to learn from a single demonstration video, but was only applied to simple tasks, such as pushing and stacking, and constrained to a 4D space (3D translation plus in-plane rotation) \cite{sieb2020graph}. In contrast, this work learns object manipulation in the full $SE(3)$ space and considers more complex and high-precision tasks, such as gear insertion and battery assembly.

\noindent\textbf{Visual Feedback Closed-Loop Manipulation} involves monitoring task state during execution and provides feedback for reactive planning to compensate execution error and scene updates. Recent work has developed closed-loop manipulation policies by integrating a 6 DoF object motion tracker and a reactive motion planner \cite{kappler2018real,andrewbwrss}. Nevertheless, the dependency on an object CAD model for tracking prevents generalization beyond a specific object. 
Given recent advances in deep reinforcement learning (RL), a number of efforts learn visuo-motor controllers by directly predicting optimal control commands from image observations \cite{levine2016end,levine2018learning,viereck2017learning,andrychowicz2020learning}, or design model predictive controllers with learned visual dynamic models \cite{ebert2018visual,manuelli2020keypoints,byravan2018se3} In contrast to these methods, the proposed closed-loop manipulation framework based on visual feedback does not require robot-object interaction for training, and can be  applied to novel objects and environments without time-consuming data collection or re-training.

\section{PROBLEM SETUP}

The input to the framework per object category $\mathcal{C}$ and associated task $\mathcal{T_C}$, e.g., {\tt Gear} as an object category and inserting a gear into a shaft as the task, is the following:
\begin{myitem}
    \item Offline: A collection $\mathbb{O}_\text{train}$ of 3D CAD object models in $\mathcal{C}$ for training, which do not include the testing objects.
    \item Demonstration: A single visual demonstration $\mathcal{D^C_T}(\mathcal{O_{D}})$ of task $\mathcal{T_C}$, i.e., an RGBD video (a gray scale and depth video are used in the accompanying experiments) recording the task execution trajectory using one of the training objects $\mathcal{O_{D}}\in \mathbb{O}_\text{train}$. $\mathcal{D^C_T}$ can be a third-person view of a tele-operated robot, or of a human performing the task. 
    \item Online: RGBD images $\mathcal{I}_t$ streamed from a camera during the execution stage when manipulating a new object instance within $\mc{C}$. 
\end{myitem}


The objective is to master the manipulation skill from the single demonstration so it can be readily applied to unseen objects in the same category $\mathcal{T_C}$ without additional fine-tuning or robot-object interactions. In addition, the manipulation skills considered here may require a sequence of actions to be executed, for which solely specifying the target configuration is not sufficient.


\section{APPROACH}

\begin{figure*}
\centering
\includegraphics[width = 0.95\textwidth]{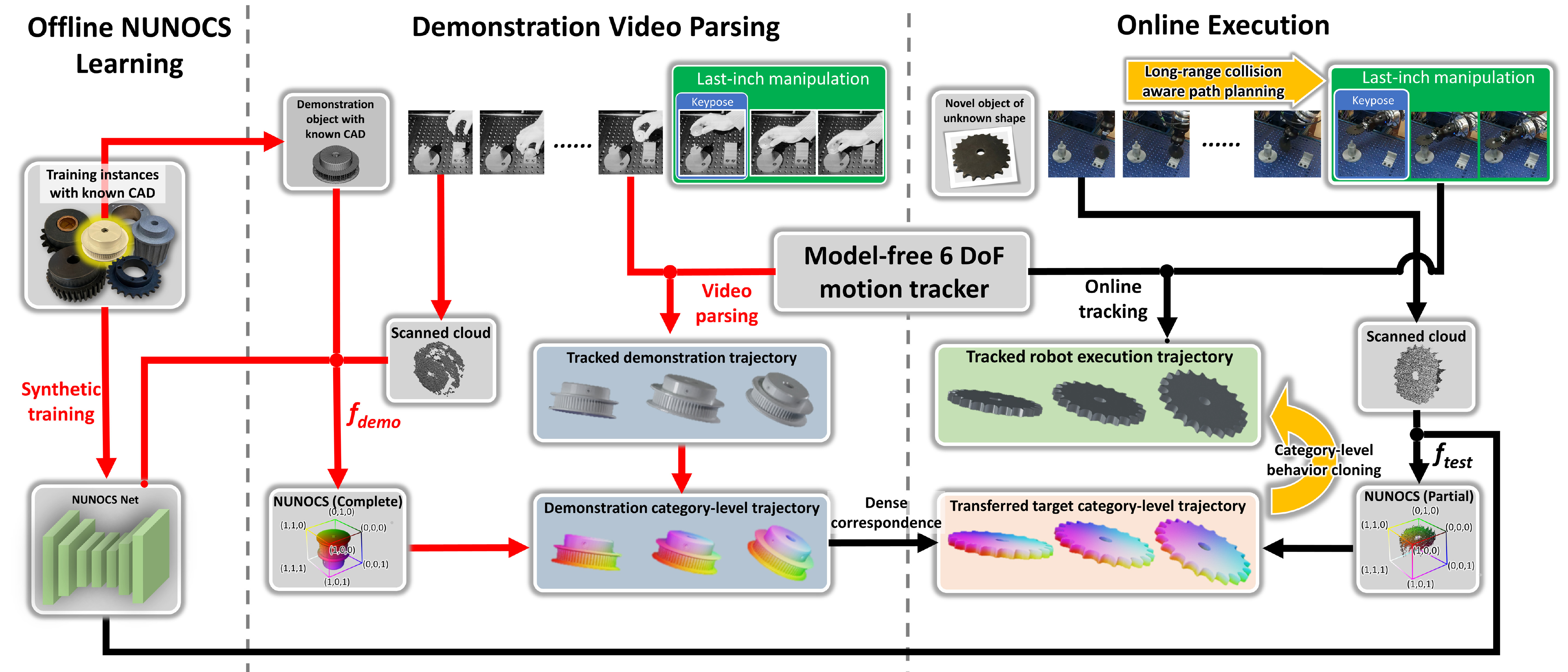}
\vspace{-.1in}
\caption{{\bf During offline NUNOCS learning}, the NUNOCS Net is trained using synthetic data generated using the training CAD models $\mathbb{O}_\text{train}$. The purpose of NUNOCS Net is to map an input point cloud to the Non-Uniform Normalized Object Coordinate Space (NUNOCS) for the object category, from which a 9D pose (translation, rotation and 3D scaling) of the observed instance in the category canonical frame can be solved in closed-form \cite{umeyama1991least}. {\bf Upon demonstration}, a model-free 6 DoF motion tracker parses the video and tracks the trajectory of the demonstrated object $\mc{O_D}$. This tracked trajectory is then lifted to a category-level demonstration trajectory by using the NUNOCS representation. In particular, the NUNOCS Net predicts the category-level object pose of the demonstrated object $\mc{O_D}$ in the first video frame. Given the 3D model of $\mc{O_D}$ and the category-level pose, a mapping $f_{\text{demo}}$ between the NUNOCS shape and the scanned cloud of $\mc{O_D}$ in the first video frame is obtained. {\bf During testing} on a novel object $O$, the NUNOCS Net takes the scanned cloud and predicts the mapping $f_\text{test}$ to its category-level NUNOCS representation. It then establishes a dense correspondence between the NUNOCS representation of $O$ and the NUNOCS shape of $\mc{O_D}$ by finding nearest neighbors, which enables to transfer the demonstration category-level trajectory to a new trajectory tailored for the target novel object $O$. The manipulation process is split into long-range, collision-free motion and last-inch manipulation. For the latter part, category-level behavior cloning is employed, which aims to clone the target category-level trajectory. Visual feedback for this process is provided by a 6 DoF motion tracker and allows behavioral cloning to adapt the manipulation of the object so that it closely follows the target trajectory until task completion.
\textcolor{red}{Red} arrows and  text  denote data flow that occurs exclusively offline.
}
\vspace{-.2in}
\label{fig:pipeline} 
\end{figure*}
 

Fig. \ref{fig:pipeline} provides an overview of the proposed framework. For each demonstration video frame, the object state is extracted via a model-free 6 DoF motion tracker \cite{wen2021bundletrack}. This allows to represent the task demonstration with an extracted trajectory $\mathcal{J^C_T}:=\{\xi_{0},\xi_{1},...,\xi_{t}\}$, where $\xi\in SE(3)$ denotes the object pose at a given timestamp.
Object poses are expressed in the receptacle's coordinate frame (e.g., the gear's pose relative to the shaft in the gear insertion task), which allows generalization to new scene configurations regardless of absolute poses.


Given the object pose trajectory parsed from the single visual demonstration, the goal is to  ``reproject'' this trajectory to other object instances in the same category. To this end, this work proposes \textbf{category-level behavior cloning (CatBC)}, which follows a virtual target pose trajectory tailored for a novel instance $\mc{O}$, reprojected from the demonstrated trajectory $\mathcal{J^C_T}$. Specifically, dense correspondence between $\mathcal{O_{D}}$ and $\mc{O}$ can be established via a category-level canonical space representation, and consequently their relative transformation can be computed. Once the virtual target trajectory for object $\mc{O}$ is obtained, behavior cloning reduces to path following by comparing the tracked pose with the reference pose.


The original demonstration video starts before $\mathcal{O_{D}}$ is grasped. The initial image frame is used to estimate the category-level pose to initialize the 6 DoF motion tracker. The ``last-inch'' action sequence is the crucial part of the manipulation process for task success.  Consider a concrete example where a gear is grasped without its opening being obstructed. In order to insert the gear into the shaft, it is the final part of the demonstration when the gear is close to the shaft that encodes the relevant spatial relation sequence between the gear and the shaft. Loosely speaking, this spatial relation sequence defines an effective manipulation policy leading to task success. Inspired by this observation, this work identifies first a \textit{keypose} as the pose that corresponds to the start of the ``last-inch'' demonstration trajectory, and marks the beginning of the category-level behavior cloning process. During the testing stage, a robot path planner is adopted to find a collision-free path that brings the manipulated object to the \textit{keypose}. This step is followed by the category-level behavior cloning for last-inch manipulation until task accomplishment. This work assumes the robot grasps the object in a way that doesn't obstruct the downstream manipulation task.


\subsection{Offline Learning of a Category-Level Representation}
\label{sec:nunocs}

Given the single visual demonstration for object $\mc{O_D}\in \mathbb{O}_\text{train}$ and in order to ``project'' the trajectory so it works for a novel object $\mc{O}$ during online execution, category-level data association between $\mc{O_D}$ and $\mc{O}$ is required. To do so, this work establishes dense correspondence in a 9-dim. space, which refers to a 6D pose and 3D scaling, to relate $\mathcal{O_D}$ to an arbitrary object instance $\mc{O}$ in the same category.  The 9-dim. space is an extension of the Normalized Object Coordinate Space (NOCS) \cite{wang2019normalized} developed for category-level 6D pose and 1D uniform scale estimation. The 9-dim. extension, referred to as "Non-Uniform Normalized Object Coordinate Space" (NUNOCS), allows for 3D scaling and has been used before for category-level task-relevant grasp planning \cite{wen2021catgrasp}. This work adopts the NUNOCS representation for complex, longer horizon tasks and category-level behavior cloning.

Concretely, given the training object models, the category-level NUNOCS representation is obtained by normalizing the corresponding point clouds along each dimension to reside within a unit cube space:
\vspace{-.05in}
$$\mc{O}_\mathbb{C}=(p-p_{min})/(p_{max}-p_{min}), \forall p \in P_\mc{O}, \mc{O}\in \mathbb{O}_\text{train},\vspace{-.05in}$$
where $p$ is a 3D point from the object point cloud, $\mathbb{C}$ denotes the canonical unit cube space shared among all objects within the same category. For an arbitrary, unknown instance $\mc{O}$, if its NUNOCS representation is available, its relationship with the known object set $\mathbb{O}_\text{train}$ can be established. 


During online execution, however, only a scanned partial point cloud of the object $\mathcal{P}_\mc{O} \in R^{N\times 3}$ is available, preventing the above operation from being applied directly. To address this issue, this work constructs a neural network, referred to here as the NUNOCS Net, to learn a mapping from a scanned partial point cloud of an instance to its configuration in the canonical unit cube space of NUNOCS, i.e. $\Phi(\mathcal{P}_\mc{O})=\mc{P}_\mathbb{C} \in R^{N\times 3}$. The mapping $\Phi$ is built with a PointNet-like architecture \cite{qi2017pointnet}. Different from \cite{wen2021catgrasp}, this work uses a separate branch to predict 3D non-uniform scales simultaneously with the input cloud's point-wise coordinates in the NUNOCS, i.e., $\Phi(\mathcal{P}_\mc{O})=(\mc{P}_\mathbb{C}, s)$ where $s=(1,\alpha,\beta)^T\in R^3$. The 3D scales $s$ are normalized w.r.t. the first dimension for compactness. During online execution, the predicted non-uniform scaling is first applied to the predicted NUNOCS coordinates as $s \circ \mc{P}_\mathbb{C}$. Subsequently, the 7D uniform scaled transformation between $s \circ \mc{P}_\mathbb{C}$ and $\mathcal{P}_\mc{O}$ is solved in closed form using least-squares \cite{umeyama1991least}, which circumvents exhaustive RANSAC iterations for solving 9D transformation in \cite{wen2021catgrasp}. Hence, the training loss is the weighted sum of the NUNOCS loss and the scaling loss:
\vspace{-.05in}
\begin{equation}
\mc{L}=\lambda_1\mc{L}_\text{NUNOCS}+\lambda_2\mc{L}_s ,
\end{equation}
\vspace{-.15in}
\begin{equation}
\mc{L}_\text{NUNOCS}=\min_{Q\in \mathbb{Q}}\sum_{p=1}^{N}\sum_{b=1}^{B}-\overline{\mc{P}}_\mathbb{C}^{(p,b)}\log(Q\mc{P}_\mathbb{C}^{(p,b)}) ,\\
\end{equation}
\vspace{-.15in}
\begin{equation}
\mc{L}_s=\left \| s-\overline{s} \right \|_2,
\vspace{-.05in}
\end{equation}
where $\overline{s}$ and $\overline{\mc{P}}_\mathbb{C}$ are the ground-truth labels. The terms $\lambda_1$ and $\lambda_2$ are the balancing weights and are empirically set to 1 in all experiments. The NUNOCS representation learning with $\mc{L}_\text{NUNOCS}$ is formulated as a classification problem by discretizing each coordinate dimension into $B$ bins ($B=100$ in all experiments) for one-hot vector encoding.
This classification formulation with a cross-entropy loss is more effective than regression as it reduces the continuous solution space to a finite number of bins $B$ \cite{wang2019normalized}. To handle symmetrical objects, $Q \in \mathbb{Q}$ are the equivalent symmetric transformations \cite{wang2019normalized}, which are pre-defined for each category. For learning the non-uniform scale mapping with $\mc{L}_s$, the $L_2$ loss is adopted. 

The NUNOCS Net is trained solely with simulated data and then directly applied to the real world without any retraining or fine-tuning. To achieve this, a synthetic training data generation process is developed using Blender \cite{blender2018} and the details are introduced in the appendix. In order to bridge the sim-to-real domain gap, domain randomization \cite{tobin2017domain} is employed by extensively randomizing the object instance types, physical parameters, object's initial poses, and the table height. In addition, the  bidirectional alignment technique over depth modality \cite{wense3tracknet} is employed to reduce the discrepancy between the simulated and real world depth data. Compared to alternatives \cite{gao2021kpam,manuelli2019kpam,florence2018dense}, the NUNOCS learning process dramatically reduces human effort by avoiding real-world data collection and additional manual annotation of keypoints \cite{gao2021kpam,manuelli2019kpam}. Dense point-wise correspondence inherited from NUNOCS also circumvents the trouble of defining the number of semantic keypoints and their locations for each category or task. While there is prior work that builds upon dense correspondence \cite{florence2018dense}, matching points over 2D image pairs tends to suffer from view ambiguity and occlusions, as validated in the experiments. 


\subsection{Model-free 6 DoF Object Motion Tracking}
\label{sec:tracking}

This work utilizes 6 DoF motion tracking for 2 purposes. During the demonstration phase, it parses the recorded video to extract the 6 DoF motion trajectory of the manipulated object in the receptacle's coordinate frame. Compared to learning directly in the image pixel space \cite{mandlekar2020learning}, this approach disentangles the object of interest from the background and represents the extracted trajectory independent of any specific scene configuration. This enables the representation to generalize to novel environments, where the initial object and receptacle placement might differ from the demonstration.




During online execution, motion tracking provides visual feedback for closed-loop control when manipulating a testing object. Uncertainty due to dynamics is unavoidable in manipulation, such as unsynchronized finger touching during grasping, in-hand object slipping and object motion caused by contacts with the receptacle during last inch manipulation. In the context of high-precision manipulation tasks, as in this work, the uncertainty introduces non-negligible errors and complicates the process of following the nominal trajectory. 


For the above purposes, this work leverages an RGBD-based 6 DoF object motion tracker \texttt{BundleTrack} \cite{wen2021bundletrack}, which provides near real-time feedback to guide the execution by comparing the estimated object pose at each timestamp against the demonstrated nominal trajectory. Alternative 6 DoF object motion trackers that rely on object CAD models \cite{wense3tracknet, deng2021poserbpf, wang20196-pack} would impede instant deployment to novel objects.


For initialization, the tracker takes as input the binary segmentation mask indicating the foreground region of interest. In particular, it reuses the inferred object mask also used by the NUNOCS Net (Sec. \ref{sec:nunocs}). The segmentation is computed by background point cloud subtraction and plane removal, followed by DBSCAN clustering \cite{ester1996density}. Alternative learning-based segmentation methods on 2D image \cite{chen2017deeplab,he2017mask} or 3D point cloud \cite{hu2020randla} can also be used. At each timestamp  $\tau \in \{1,2,...,t\}$, the process tracks the object motion relative to the initial timestamp in the camera's frame, $\xi_{0 \rightarrow \tau} \in SE(3)$. To obtain the absolute category-level pose in the camera frame at $\tau$, a transformation is applied to the initial category-level pose $\xi_{0}$ inferred by the NUNOCS Net, i.e. $\xi_{\tau} = \xi_{0}[(\xi_{0})^{-1} \xi_{0 \rightarrow \tau} \xi_{0}] = \xi_{0 \rightarrow \tau} \xi_{0} \in SE(3)$. As the statically-mounted camera has been calibrated relative to the robot base, the poses $\xi_{\tau}$ can be further expressed in the robot frame so as to be used by other modules. During the demonstration, the receptacle pose is estimated using a model-based pose estimation approach \cite{mitash2020scene}. In this way, the demonstrated object pose trajectory can be represented in the receptacle's coordinate frame, enabling behavior cloning in different environment configurations.

Once initialized, the method is able to track the object at the RGBD camera's acquisition rate (10 Hz) without any re-initialization. The neural network weights in the tracker are adopted from the available, open-sourced implementation\footnote{\url{https://github.com/wenbowen123/BundleTrack}} and remain fixed in all experiments, eliminating the requirement of training data collection. Since the tracker does not require CAD models for the objects, it can be instantly applied to arbitrary objects in both scenarios, i.e., demonstration video parsing and online visual feedback control. 


\begin{figure*}[h]
\centering
\includegraphics[width = 0.95\textwidth]{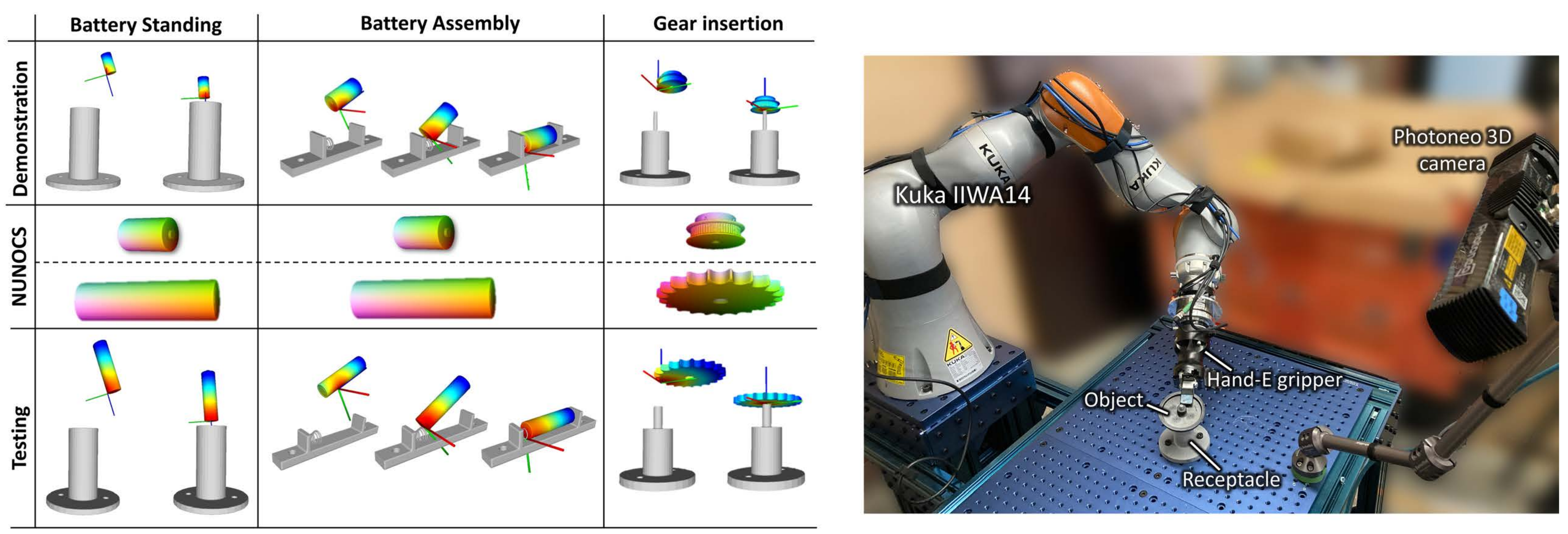}
\vspace{-0.15in}
\caption{\textbf{Left:} Heatmap visualizations of the local attention mechanism are shown in the top and bottom rows for the training and testing objects respectively. During demonstration, given the 3D model of the demonstration object and its paired receptacle, an attention heatmap is precomputed. During online execution, the attention heatmap can be transferred to a novel object given the dense correspondence established through their NUNOCS representations shown in the middle row. The attention mechanism allows to dynamically anchor the coordinate system to the local attended point (located at the warmest color), capturing the variation between demonstration and testing objects in scaling and local typology. The testing objects' 3D models are shown for visualization only and are unknown during execution. \textbf{Right:} Experimental hardware setup.}
\label{fig:attention} 
\vspace{-0.25in}
\end{figure*}


\subsection{Category-Level Behavior Cloning as Last-Inch Policy}\label{sec:catbc}

\vspace{-0.1in}

    	    


\vspace{-0.05in}
\begin{algorithm}[h]
\label{alg:cbc}
\caption{Category-Level Behavior Cloning}
\KwIn{\textbf{tracker}, \textbf{robot}, $\mc{J}$ \tcp{virtual target trajectory}} 
\tcp{starting from the \textit{keypose}}
\For{$\overline{\xi}_i$ in $\mc{J}$}
{ 
    
    $\xi \gets \textbf{tracker}.get\_pose()$ \tcp{object  $SE(3)$ pose}
    $\Delta\xi \gets \overline{\xi}_i \ominus \xi$ \tcp{relative pose}
    $q \gets $ \textbf{robot}.$get\_joints()$ \tcp{joint configuration} 
    $J \gets \textbf{robot}.get\_jacobian(q)$ \tcp{Jacobian matrix}
    $\Delta q \gets J^{\dagger} \Delta\xi$ \tcp{Jacobian steering}
    $q' \gets q+\Delta q$ \\
    \textbf{robot}.$reach(q')$ \tcp{move joints}
}
\end{algorithm}
\vspace{-0.05in}

\vspace{-0.1in}

Once the object pose trajectory is obtained by parsing the demonstration video (Sec. \ref{sec:tracking}), a canonicalization in the NUNOCS space is performed (Sec. \ref{sec:nunocs}) to establish dense correspondence between the demonstrated object $\mc{O_D}$ and the testing object $\mc{O}$, as shown in Fig. \ref{fig:pipeline}. This effectively ``reprojects'' the demonstration trajectory to a virtual target trajectory tailored for $\mc{O}$. This allow to replay the actions performed in the demonstration so as to accomplish the task with the novel instance $\mc{O}$. Specifically, even without prior knowledge about $\mc{O}$, following the target trajectory allows task success. This is because by following the virtual target trajectory, the novel object $\mc{O}$ traverses the critical task-relevant configurations relative to the receptacle in a desired sequence. This process is referred to here as \textbf{Category-level Behavior Cloning (CatBC)}. CatBC realizes a manipulation policy by replicating the demonstration trajectory, which is defined in an object-centric manner and is agnostic to how the object is grasped by the robot. To increase the robustness of CatBC against uncertainties due to manipulation dynamics, constantly updated object state is needed to ensure the object follows the target path. In contrast, previous treatments of the grasped object as an extended kinematic frame \cite{gao2021kpam,manuelli2019kpam,gualtieri2021robotic} tend to be brittle, as shown in the accompanying experiments (Sec.~\ref{sec:exp}).

Alg. \ref{alg:cbc} outlines the CatBC process. A model-free 6 DoF object motion tracker (Sec. \ref{sec:tracking}) provides online visual feedback for closed-loop control. During the last-inch manipulation, dynamics uncertainty arising from contacts and robot-object-receptacle interaction causes the behavior cloning process to deviate from the desired trajectory. It is thus necessary to discretize the trajectory densely (the neighboring poses' distance is around $2mm$ or $2\degree$ in our implementation) so that the visual feedback ensures the target trajectory is followed to the highest degree when moving to the next immediate goal along the trajectory. 

\subsection{Dynamic Category-Level Frame via Local Attention}

Typically, 6D object poses are used to represent a manipulated object's state in a predefined local coordinate system and define the transformation relative to a task-relevant target frame. Given a 6D pose and the 3D model of an object instance, any point on the rigid object is always uniquely defined. This allows to implicitly define the object's  parts and orientations relevant to a specific task. Nevertheless, it is challenging to adopt one constant, category-level, canonical coordinate frame for different tasks while capturing the geometric variation across all instances.



Consider the \textit{Batteries} class as an example. If the commonly selected center-of-mass is used as the canonical coordinate frame origin, when aligning a novel battery instance to the demonstrated one, it may collide with or float away from the receptacle, depending on its particular larger or smaller diameter. Instead, the surface center of one of the terminal ends (e.g., the negative pole of the battery) is more appropriate as the frame origin for the battery standing task. In contrast, the negative pole's lowest edge center is more appropriate as the frame origin for the battery assembly task, which comes in contact with both the receptacle and the spring. Nevertheless, it is cumbersome to manually specify a suitable local frame for each task. Moreover, the task-relevant local frame may not stay constant throughout a complex task.

This work proposes a local attention mechanism to automatically and dynamically select an anchor point $p^*_\tau$ that defines the origin of the category-level canonical coordinate system, as in Fig. \ref{fig:attention}. Concretely, during the demonstration, a signed distance function (positive external to the object's surface) of the receptacle is computed, noted as $\Omega(\cdot)$ \cite{malladi1995shape}. Then, an attention heatmap and the anchor point at any timestamp along the manipulation horizon $\tau \in \{0,1,...,t\}$ are computed as:
\vspace{-.05in}
\begin{equation}
Attn_\tau(p_i)= 1-\frac{exp(\Omega(\xi_\tau p_i))}{\sum_j exp(\Omega(\xi_\tau p_j))},
\end{equation}
\vspace{-.15in}
\begin{equation}
p^*_\tau=\argmaxB_{p_i}{Attn_\tau(p_i)},
\vspace{-.1in}
\end{equation}
\noindent where $p_i$ are the points on the 3D model of $\mc{O_D}$. $\xi_{\tau}$ denotes the demonstration object's pose relative to the receptacle along the trajectory $\mathcal{J^C_T}:=\{\xi_{0},\xi_{1},...,\xi_{t}\}$, which is parsed from the demonstration video. Intuitively, the local object part that is closer to the receptacle should be assigned higher attention. During online execution, however, the novel object's shape is not available to directly compute the attention heatmap. By virtue of the established dense correspondence using NUNOCS (Fig. \ref{fig:attention} middle row), the attention heatmap can be transferred from $\mc{O_D}$ to novel objects (Fig. \ref{fig:attention} last row). The attention mechanism allows to dynamically anchor the coordinate system to the local attended point, capturing the variation between demonstration and testing objects in scaling and local typology. The coordinate system is only translated to attend to the task-relevant local region, while the orientation remains the same as the original learned category-level canonical frame.

Compared to using a small number of pre-specified keypoints as in previous work \cite{gao2021kpam,manuelli2019kpam}, the proposed dynamic attention mechanism allows for versatile implicit keypoint generation augmented with orientation information, which self-adjusts along the manipulation horizon. This improves expressiveness, reduces human effort, and enables high precision category-level behavior cloning. 


\subsection{Grasping the Object and Transferring it to the Keypose}

The proposed framework is not constrained to a particular grasp planning approach and in general, any CAD model-free grasping planning method \cite{mahler2017dex,ten2017grasp} can be adopted. As long as the grasp complies with the downstream manipulation task, as in the considered setup, task-relevant grasp planning can be adopted. The core idea is to utilize the category-level affordance priors unified by NUNOCS representations to propose and select task-relevant grasps \cite{wen2021catgrasp}. Grasp planning is not the focus of this work.

The proposed framework is robust to uncertainty due to robot-object interactions, such as the object moving during finger closing. This is achieved due to the online object state update from the 6 DoF tracker. Once the object is grasped, the tracker provides the latest in-hand object pose, which serves as the start pose for a path planner (RRT* \cite{karaman2011sampling} in the implementation) to find a collision-free path that transports the object to the \textit{keypose}. The decomposition into long-range, collision-free motion to bring the object to the \textit{keypose}, and then, last-inch manipulation, provides an implicit attention mechanism that ignores the unimportant background information in the demonstration. It focuses on the critical actions that define task accomplishment. The long-range collision-aware path planning also ensures safety when applying the framework to new scenes with novel obstacles (Sec. \ref{sec:analysis}). The choice of \textit{keypose} is insensitive and empirically set as the pose $5cm$ away from the receptacle along the demonstrated trajectory in all our experiments.


\section{EXPERIMENTS}\label{sec:exp}

\subsection{Experimental Setup}


The evaluation is performed exclusively with real-world experiments. The hardware is composed of a Kuka IIWA14 arm, a Robotiq Hand-E gripper, a Photoneo 3D camera providing gray scale and depth images at 10 Hz, as well as a spring-damper device mounted between the gripper and the robot flange that  provides passive compliance (see Fig. \ref{fig:attention}). For better accessibility, the robot is controlled in position mode with the joint position commands computed by manipulation policies. 
Computations are  conducted on a standard desktop with an Intel Core i9-10900X CPU processor and a single NVIDIA RTX 2080 Ti GPU for both training and testing. 

\begin{figure}[h]
\centering
\includegraphics[width = 0.45\textwidth]{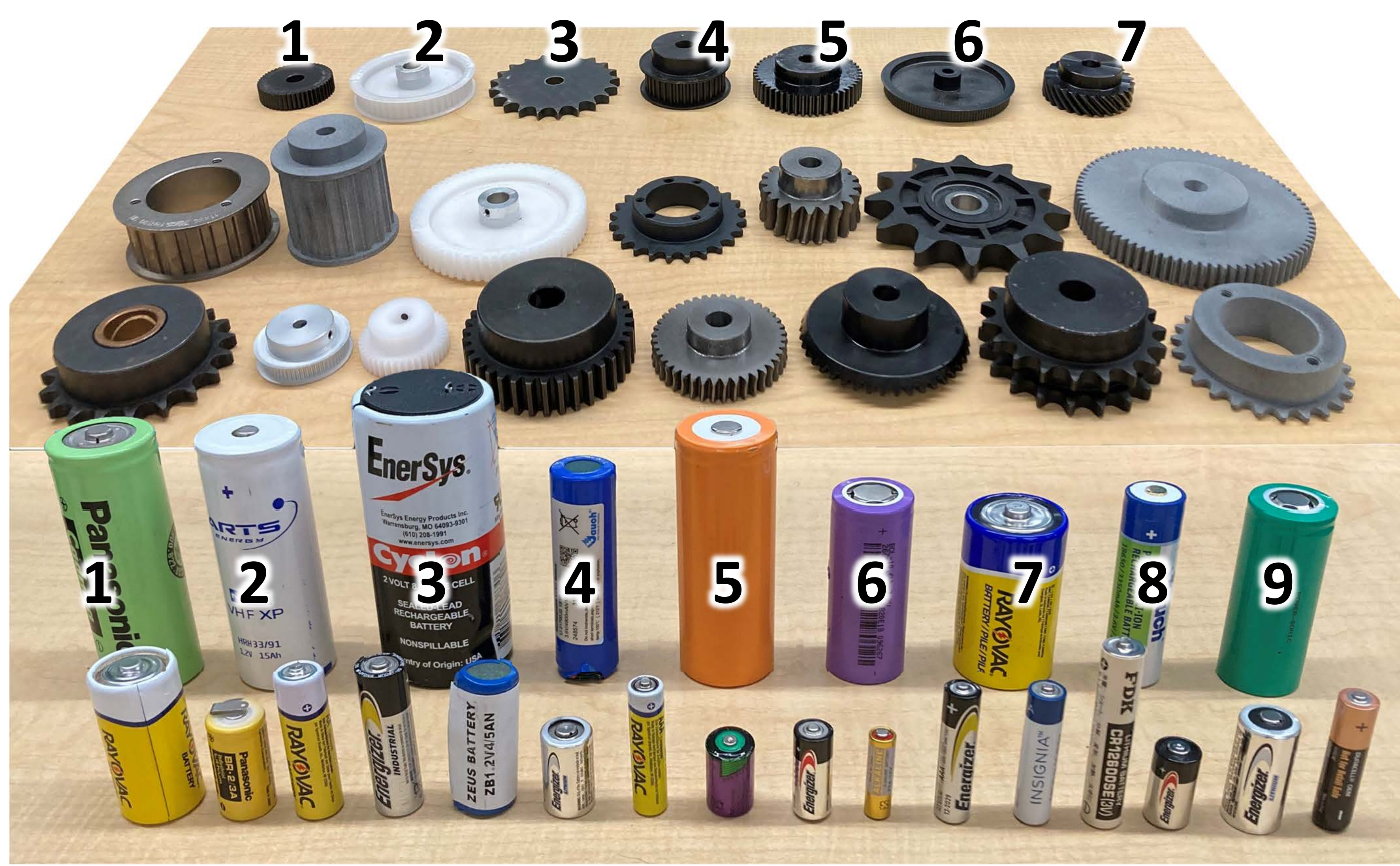}
\vspace{-0.1in}
\caption{Experimental objects in categories \textit{Gears} and \textit{Batteries}. In each category, the testing object set are labeled with IDs. The rest are the training object instances with known CAD models. Note that the real world training objects are only used for data collection to train baseline methods. The proposed approach learns solely in simulation using their CAD models. The testing objects are selected to be manipulable with the gripper but otherwise diverse in shape and appearance cross instances.}
\label{fig:objects} 
\vspace{-.2in}
\end{figure}

There are 2 object categories considered: \textit{Gears} and  \textit{Batteries}. The training and testing splits are depicted in Fig. \ref{fig:objects}. The real world training objects are used for data collection to train baseline methods while the proposed approach only requires virtual 3D models to learn in simulation. The test set, 7 instances for \textit{Gears} and 9 for \textit{Batteries}, are real industrial or commercial objects purchased from popular retailers. They are chosen to vary in shape and appearance to evaluate the cross-instance generalization of the methods. They are different from the training objects.



Three manipulation tasks (Sec. \ref{sec:battery_standing_task}, \ref{sec:battery_task}, and \ref{sec:gear_task}) are defined with different complexity and tolerance levels. For each task, a single video demonstration of a human manipulating a randomly selected training object is recorded offline. This single demonstration is utilized to perform manipulation of novel objects with no additional data collection. The experiment in each object-task-tolerance setting consists of 5 trials unless otherwise specified, with different initial configurations for both the object and the receptacle. The receptacles' shapes are designed from geometric primitives and their configurations during testing are detected using RANSAC \cite{fischler1981random}. A total number of 1560 task trials were executed with the proposed method and 7 baselines (Sec.~\ref{sec:baseline}). The following questions are explored: 1) How well are the manipulation skills learned from a single video demonstration? 2) How well do they generalize to novel instances? 3) How robust are the skills to different scenes and uncertainty due to dynamics? 4) What level of contact-rich interaction can the learned policy achieve?



\subsection{Baseline Methods}
\label{sec:baseline}
Comparison points correspond to state-of-the-art, vision-based, category-level manipulation methods. The baselines have been tuned for improved performance. Details are provided in the Appendix.

\textbf{DON}\cite{florence2018dense}, \textbf{KPAM}\cite{manuelli2019kpam}, and \textbf{KPAM 2.0} \cite{gao2021kpam}: These 3 methods are based on open-source implementations. The original training data collection pipeline is applied with the training objects in this work. The tasks are defined by manually specifying the task-relevant keypoints. The Appendix provides example training data and annotated keypoints.

\textbf{DON BC}, \textbf{KPAM BC}, and \textbf{KPAM 2.0 BC}: These 3 methods are Behavior Cloning \textbf{(BC)}-augmented versions, where the manual goal specification is replaced by visual demonstration. The proposed video parsing and CatBC framework are integrated with the prior methods to capture the intermediate action sequence and realize a closed-loop policy.

\textbf{Ours-no-tracking}: In order to study the effectiveness of visual feedback control for handling uncertainty due to manipulation dynamics, this variation performs open-loop control by disabling the visual motion tracker.

\subsection{Battery Standing Task} \label{sec:battery_standing_task}

\noindent \textbf{Setup:} This task requires the robot to grasp the battery from the table-top and place it vertically on a small cylindrical platform, such that the battery stands stably after being released from the gripper. The demonstration video and an example successful robot execution are shown in the first page's figure (top). This task represents commonly considered pick-and-place tasks in prior efforts on visual imitation learning \cite{huang2019neural, manuelli2019kpam, jin2020visual, sieb2020graph, xiong2021learning, mandlekar2020learning, tremblay2018synthetically,liang2022learning}. This task is the simplest among the ones considered here but still requires accurate orientation reasoning for stable placement due to the batteries' shape, which is  usually long and thin.


\vspace{-.1in}
\begin{table}[h]
\centering
\def\mywidth{0.45\textwidth}
\rule{\mywidth}{2pt}
\resizebox{\mywidth}{!}{
\begin{tabular}{c|cccccccc}
\multirow{2}[1]{*}{Instance} & KPAM  & KPAM  & KPAM  & KPAM  & DON   & DON   & Ours-no & \multirow{2}[1]{*}{Ours} \\
      & \cite{manuelli2019kpam} & BC \cite{manuelli2019kpam} & 2.0 \cite{gao2021kpam} & 2.0 BC \cite{gao2021kpam} & \cite{florence2018dense} & BC \cite{florence2018dense} & tracking &  \bigstrut[b]\\
\hline
battery1 & 3/5   & 3/5   & 5/5   & 5/5   & 1/5   & 2/5   & 5/5   & \cellcolor[rgb]{ .847,  .847,  .847}5/5 \bigstrut[t]\\
battery2 & 5/5   & 5/5   & 4/5   & 4/5   & 2/5   & 1/5   & 5/5   & \cellcolor[rgb]{ .847,  .847,  .847}5/5 \\
battery3 & 1/5   & 1/5   & 1/5   & 2/5   & 3/5   & 2/5   & 5/5   & \cellcolor[rgb]{ .847,  .847,  .847}5/5 \\
battery4 & 1/5   & 0/5   & 0/5   & 0/5   & 0/5   & 0/5   & 2/5   & \cellcolor[rgb]{ .847,  .847,  .847}5/5 \\
battery5 & 5/5   & 5/5   & 5/5   & 5/5   & 0/5   & 0/5   & 5/5   & \cellcolor[rgb]{ .847,  .847,  .847}5/5 \\
battery6 & 1/5   & 1/5   & 2/5   & 2/5   & 0/5   & 0/5   & 3/5   & \cellcolor[rgb]{ .847,  .847,  .847}5/5 \\
battery7 & 3/5   & 3/5   & 3/5   & 5/5   & 2/5   & 1/5   & 5/5   & \cellcolor[rgb]{ .847,  .847,  .847}5/5 \\
battery8 & 4/5   & 5/5   & 4/5   & 3/5   & 0/5   & 0/5   & 2/5   & \cellcolor[rgb]{ .847,  .847,  .847}5/5 \\
battery9 & 0/5   & 0/5   & 2/5   & 1/5   & 0/5   & 0/5   & 4/5   & \cellcolor[rgb]{ .847,  .847,  .847}5/5 \bigstrut[b]\\
\hline
Total & 51.1\% & 51.1\% & 57.8\% & 60.0\% & 17.8\% & 13.3\% & 80.0\% & \cellcolor[rgb]{ .847,  .847,  .847}\textbf{100.0\%} \bigstrut[t]\\
\end{tabular}%
}
\rule{\mywidth}{2pt}
\vspace{-0.1in}
\caption{Results of battery standing task. The testing instances are shown in Fig. \ref{fig:objects}.}
\label{tab:standing}
\end{table}
\vspace{-0.15in}

\vspace{-.1in}

\noindent \textbf{Results:} The quantitative comparison is presented in Table \ref{tab:standing}. For \textbf{KPAM}, \textbf{KPAM 2.0}, and \textbf{DON}, common failure cases arise due to the detected keypoints or dense correspondence not being able to provide reliable constraints for goal specification, i.e., the battery standing vertically.  The performance gain by behavior cloning \textbf{(BC)} is insignificant. The reason is the exact last-inch action sequence is not crucial in this pick-and-place task; namely, the task succeeds as long as collision can be avoided while moving the battery towards the correct goal configuration. \textbf{Ours} consistently accomplishes the task regardless of the battery instance. Nevertheless, when motion tracking is disabled, the performance drops to $80.0\%$ as the gripper-battery interaction during the gripper's closing perturbs the battery to a different orientation from the initial estimate, thus leading to inclined and unstable battery placement. When compared to \textbf{KPAM BC}, \textbf{KPAM 2.0 BC} and \textbf{DON BC}, \textbf{Ours-no-tracking} yields a higher success rate, indicating that the NUNOCS representation achieves more accurate characterization of the objects' poses and improves cross-instance generalizability.

\subsection{Battery Assembly Task} \label{sec:battery_task}


\noindent \textbf{Setup:} This task requires the robot to pick the battery and insert it into a receptacle where a spring is mounted on the internal side of a vertical wall.  The receptacles are designed with their lengths and wall heights matching their paired battery's dimension. The spring must be pressed to at least 1/2 of its original length (16 mm) to reserve enough space for the battery. To accomplish the task, the battery has to first press the spring with its negative terminal end to reserve enough space for the positive end to be pressed down. Finally, the gripper releasing the battery allows the spring to stretch, pushing the battery tightly towards the other wall to stay stably inside the receptacle. The experiments evaluate the steady state and mark the cases as failed when the spring is squeezed sideways due to brute force from the gripper. The demonstration video and an example successful robot execution are illustrated in the first page's figure (middle). This task requires robustness against uncertainty due to rich contact and external forces from the environment. It highlights the challenges in learning long-horizon manipulation policies where the last-inch action sequence is critical to task success.



\vspace{-.1in}
\begin{table}[h]
\centering
\def\mywidth{0.45\textwidth}
\rule{\mywidth}{2pt}
\resizebox{\mywidth}{!}{
\begin{tabular}{c|cccccccc}
\multirow{2}[1]{*}{Instance} & KPAM  & KPAM  & KPAM  & KPAM  & DON   & DON   & Ours-no & \multirow{2}[1]{*}{Ours} \\
      & \cite{manuelli2019kpam} & BC \cite{manuelli2019kpam} & 2.0 \cite{gao2021kpam} & 2.0 BC \cite{gao2021kpam} & \cite{florence2018dense} & BC \cite{florence2018dense} & tracking &  \bigstrut[b]\\
\hline
battery1 & 0/5   & 2/5   & 0/5   & 2/5   & 0/5   & 0/5   & 2/5   & \cellcolor[rgb]{ .847,  .847,  .847}5/5 \bigstrut[t]\\
battery2 & 0/5   & 1/5   & 0/5   & 1/5   & 0/5   & 2/5   & 1/5   & \cellcolor[rgb]{ .847,  .847,  .847}3/5 \\
battery3 & 0/5   & 2/5   & 0/5   & 2/5   & 0/5   & 2/5   & 3/5   & \cellcolor[rgb]{ .847,  .847,  .847}5/5 \\
battery4 & 0/5   & 0/5   & 0/5   & 0/5   & 0/5   & 0/5   & 0/5   & \cellcolor[rgb]{ .847,  .847,  .847}2/5 \\
battery5 & 0/5   & 1/5   & 0/5   & 1/5   & 0/5   & 0/5   & 2/5   & \cellcolor[rgb]{ .847,  .847,  .847}5/5 \\
battery6 & 0/5   & 0/5   & 0/5   & 0/5   & 0/5   & 0/5   & 2/5   & \cellcolor[rgb]{ .847,  .847,  .847}5/5 \\
battery7 & 0/5   & 2/5   & 0/5   & 2/5   & 0/5   & 1/5   & 2/5   & \cellcolor[rgb]{ .847,  .847,  .847}5/5 \\
battery8 & 0/5   & 2/5   & 0/5   & 2/5   & 0/5   & 1/5   & 3/5   & \cellcolor[rgb]{ .847,  .847,  .847}3/5 \\
battery9 & 0/5   & 0/5   & 0/5   & 1/5   & 0/5   & 0/5   & 2/5   & \cellcolor[rgb]{ .847,  .847,  .847}4/5 \bigstrut[b]\\
\hline
Total & 0.00\% & 22.22\% & 0.00\% & 24.44\% & 0.00\% & 13.33\% & 37.78\% & \cellcolor[rgb]{ .847,  .847,  .847}\textbf{82.22\%} \bigstrut[t]\\
\end{tabular}%
}
\rule{\mywidth}{2pt}
\vspace{-0.1in}
\caption{Results of battery assembly task.  The testing instances are shown in Fig. \ref{fig:objects}.}
\label{tab:batttery_assembly}
\end{table}
\vspace{-0.15in}

\vspace{-.1in}

\noindent \textbf{Results:} The quantitative comparison is presented in Table \ref{tab:batttery_assembly}. When only specifying the goal configuration and directly transporting the battery to the goal, \textbf{KPAM}, \textbf{KPAM 2.0}, and \textbf{DON} are not able to accomplish the task. The spring is often squeezed by excessive force but not along the principal axis. Therefore, upon gripper releasing, the battery usually pops out of the receptacle pushed by the spring. Augmenting the 3 baselines with \textbf{BC} enables to reason over the intermediate sequential actions and improves performance. Nevertheless, complicated by the elastic battery-spring interaction and the battery-receptacle friction, dynamics uncertainty frequently results in significant in-hand object motion, causing the open-loop execution to deviate from the desired target trajectory. This is also reflected by the large performance gap between \textbf{Ours-no-tracking} and \textbf{Ours}. In \textbf{Ours}, the motion tracker constantly provides visual feedback about the latest battery state, ensuring the closed-loop policy can guide the battery following the target category-level trajectory for task success.

\subsection{Gear Insertion Task}\label{sec:gear_task}

\noindent \textbf{Setup:} This task requires the robot to insert the gear into a tight-tolerance shaft. To investigate the precision boundary of the category-level manipulation approaches, experiments are conducted on varying levels of gear-shaft tolerances including $0.1mm$, $0.5mm$, $5mm$ (or $3mm$ if limited by the hole diameter of the gear). The shafts are designed to have similar lengths and varying diameters to realize different tolerances. A task trial is marked as success if the gear's hole passes through the shaft. This task highlights the common challenges in contact-rich manipulation: requiring high precision and robustness against uncertainty. The demonstration video and an example successful robot execution are shown in the first page's figure (bottom).

\vspace{-.1in}
\begin{table}[h]
\centering
\def\mywidth{0.45\textwidth}
\rule{\mywidth}{2pt}
\resizebox{\mywidth}{!}{
\begin{tabular}{r|c|cccccccc}
\multicolumn{1}{r|}{\multirow{2}[1]{*}{Instance}} & Tolerance & KPAM  & KPAM  & KPAM  & KPAM  & DON   & DON   & Ours-no & \multirow{2}[1]{*}{Ours} \\
      & (mm)  & \cite{manuelli2019kpam} & BC \cite{manuelli2019kpam} & 2.0 \cite{gao2021kpam} & 2.0 BC \cite{gao2021kpam} & \cite{florence2018dense} & BC \cite{florence2018dense} & tracking &  \bigstrut[b]\\
\hline
\multicolumn{1}{r|}{\multirow{3}[2]{*}{gear1}} & 0.1   & 0/5   & 0/5   & 0/5   & 0/5   & 0/5   & 0/5   & 0/5   & \cellcolor[rgb]{ .847,  .847,  .847}2/5 \bigstrut[t]\\
      & 0.5   & 0/5   & 0/5   & 0/5   & 0/5   & 0/5   & 0/5   & 1/5   & \cellcolor[rgb]{ .847,  .847,  .847}5/5 \\
      & 5     & 2/5   & 2/5   & 2/5   & 2/5   & 1/5   & 2/5   & 3/5   & \cellcolor[rgb]{ .847,  .847,  .847}5/5 \bigstrut[b]\\
\hline
\multicolumn{1}{r|}{\multirow{3}[2]{*}{gear2}} & 0.1   & 0/5   & 0/5   & 0/5   & 0/5   & 0/5   & 0/5   & 0/5   & \cellcolor[rgb]{ .847,  .847,  .847}2/5 \bigstrut[t]\\
      & 0.5   & 0/5   & 0/5   & 0/5   & 0/5   & 0/5   & 0/5   & 0/5   & \cellcolor[rgb]{ .847,  .847,  .847}5/5 \\
      & 5     & 2/5   & 2/5   & 2/5   & 2/5   & 0/5   & 0/5   & 3/5   & \cellcolor[rgb]{ .847,  .847,  .847}5/5 \bigstrut[b]\\
\hline
\multicolumn{1}{r|}{\multirow{3}[2]{*}{gear3}} & 0.1   & 0/5   & 0/5   & 0/5   & 0/5   & 0/5   & 0/5   & 0/5   & \cellcolor[rgb]{ .847,  .847,  .847}2/5 \bigstrut[t]\\
      & 0.5   & 0/5   & 0/5   & 0/5   & 0/5   & 0/5   & 0/5   & 1/5   & \cellcolor[rgb]{ .847,  .847,  .847}5/5 \\
      & 5     & 2/5   & 2/5   & 2/5   & 2/5   & 2/5   & 1/5   & 3/5   & \cellcolor[rgb]{ .847,  .847,  .847}5/5 \bigstrut[b]\\
\hline
\multicolumn{1}{r|}{\multirow{3}[2]{*}{gear4}} & 0.1   & 0/5   & 0/5   & 0/5   & 0/5   & 0/5   & 0/5   & 0/5   & \cellcolor[rgb]{ .847,  .847,  .847}1/5 \bigstrut[t]\\
      & 0.5   & 0/5   & 0/5   & 0/5   & 0/5   & 0/5   & 0/5   & 0/5   & \cellcolor[rgb]{ .847,  .847,  .847}4/5 \\
      & 5     & 1/5   & 1/5   & 1/5   & 1/5   & 0/5   & 0/5   & 2/5   & \cellcolor[rgb]{ .847,  .847,  .847}5/5 \bigstrut[b]\\
\hline
\multicolumn{1}{r|}{\multirow{3}[2]{*}{gear5}} & 0.1   & 0/5   & 0/5   & 0/5   & 0/5   & 0/5   & 0/5   & 0/5   & \cellcolor[rgb]{ .847,  .847,  .847}1/5 \bigstrut[t]\\
      & 0.5   & 0/5   & 0/5   & 0/5   & 0/5   & 0/5   & 0/5   & 0/5   & \cellcolor[rgb]{ .847,  .847,  .847}4/5 \\
      & 3     & 0/5   & 0/5   & 1/5   & 1/5   & 0/5   & 0/5   & 1/5   & \cellcolor[rgb]{ .847,  .847,  .847}5/5 \bigstrut[b]\\
\hline
\multicolumn{1}{r|}{\multirow{3}[2]{*}{gear6}} & 0.1   & 0/5   & 0/5   & 0/5   & 0/5   & 0/5   & 0/5   & 0/5   & \cellcolor[rgb]{ .847,  .847,  .847}1/5 \bigstrut[t]\\
      & 0.5   & 0/5   & 0/5   & 0/5   & 0/5   & 0/5   & 0/5   & 0/5   & \cellcolor[rgb]{ .847,  .847,  .847}2/5 \\
      & 5     & 0/5   & 0/5   & 2/5   & 2/5   & 0/5   & 0/5   & 2/5   & \cellcolor[rgb]{ .847,  .847,  .847}5/5 \bigstrut[b]\\
\hline
\multicolumn{1}{r|}{\multirow{3}[2]{*}{gear7}} & 0.1   & 0/5   & 0/5   & 0/5   & 0/5   & 0/5   & 0/5   & 0/5   & \cellcolor[rgb]{ .847,  .847,  .847}1/5 \bigstrut[t]\\
      & 0.5   & 0/5   & 0/5   & 0/5   & 0/5   & 0/5   & 0/5   & 0/5   & \cellcolor[rgb]{ .847,  .847,  .847}4/5 \\
      & 5     & 0/5   & 1/5   & 1/5   & 1/5   & 0/5   & 0/5   & 1/5   & \cellcolor[rgb]{ .847,  .847,  .847}5/5 \bigstrut[b]\\
\hline
\hline
\multicolumn{1}{r|}{\multirow{3}[1]{*}{Total}} & 0.1   & 0.0\% & 0.0\% & 0.0\% & 0.0\% & 0.0\% & 0.0\% & 0.0\% & \cellcolor[rgb]{ .847,  .847,  .847}\textbf{28.6\%} \bigstrut[t]\\
      & 0.5   & 0.0\% & 0.0\% & 0.0\% & 0.0\% & 0.0\% & 0.0\% & 5.7\% & \cellcolor[rgb]{ .847,  .847,  .847}\textbf{82.9\%} \\
      & 5     & 20.0\% & 22.9\% & 31.4\% & 31.4\% & 8.6\% & 8.6\% & 42.9\% & \cellcolor[rgb]{ .847,  .847,  .847}\textbf{100.0\%} \\
\end{tabular}%
}
\rule{\mywidth}{2pt}
\vspace{-0.1in}
\caption{Results of gear insertion task. The testing instances are shown in Fig. \ref{fig:objects}.}
\vspace{-0.15in}
\label{tab:gear}
\end{table}

\vspace{-.1in}

\begin{figure*}[t]
\centering
\includegraphics[width = 0.95\textwidth]{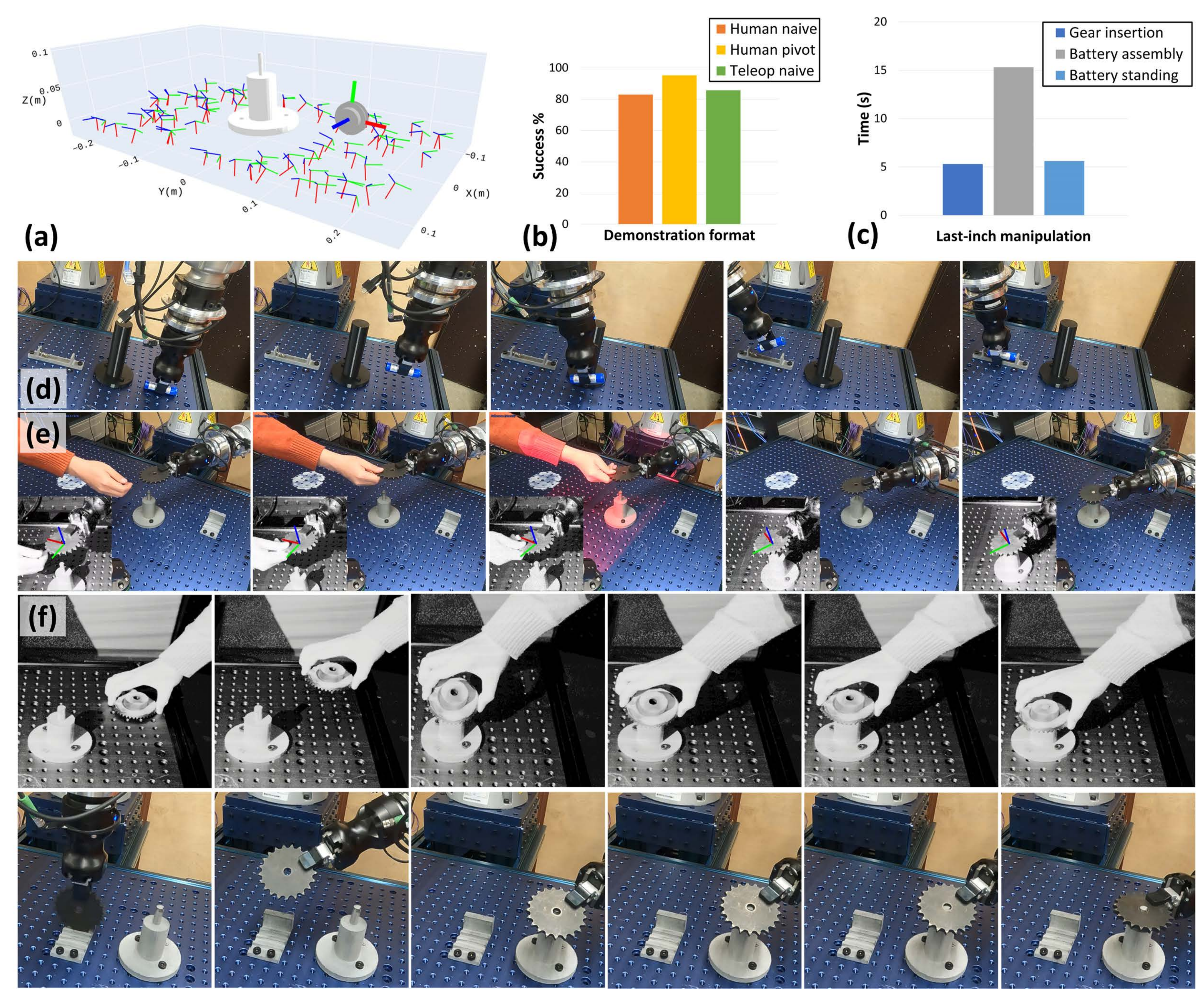}
\vspace{-0.15in}
\caption{(a) Distribution of gears' initial poses relative to the receptacle in the ``gear insertion'' experiments (Sec. \ref{sec:gear_task}). The gray gear mesh represents the demonstration object $\mc{O_D}$ in its initial configuration. During testing, the framework generalizes to unseen configurations. (b) Overall success rates of the 3 policies learned from different demonstrations in the ``gear insertion'' task. The success rates are averaged across object instances and the same number of runs for the $0.5mm$ tolerance as in Table \ref{tab:gear}. The method \textbf{Ours} in Table \ref{tab:gear} is based on ``human naive''. (c) Running time of last-inch manipulation in different tasks. (d) An example testing case of the ``battery assembly'' task, where the proposed approach generalizes to unstructured environments with obstacles unseen in the demonstration video. (e) Visual motion tracker constantly updates the object pose for robust CatBC against external disturbances, such as human dragging. Pose visualization thumbnails are in bottom-left corners. (f) For the ``gear insertion'' task, an anchor-and-pivot manipulation strategy is provided instead (first row), and the robot executes the learned policy on a testing object (second row). Complete videos are available in supplementary material.}
\vspace{-0.25in}
\label{fig:analysis} 
\end{figure*}

\noindent \textbf{Results:} The quantitative comparison is presented in Table \ref{tab:gear}. Even with the most relaxed tolerance ($5mm$), \textbf{KPAM}, \textbf{KPAM 2.0}, and \textbf{DON} struggle as the predicted semantic keypoints and cross-image 2D dense correspondences are not sufficiently reliable to achieve the required precision in the 3D space. Augmenting with \textbf{BC} delivers small improvement due to the unreliably estimated gear state. Additionally, \textit{Gears}' being textureless and reflective, poses notable challenges to the training data collection pipeline shared by  the above baselines \cite{gao2021kpam,manuelli2019kpam,florence2018dense}. In particular, a pre-determined scanning trajectory with a fixed number of view points is not able to cover the potentially novel views with different reflections. In contrast, even without visual feedback, \textbf{Ours-no-tracking} yields superior performance, validating the effectiveness of the robust and reliable NUNOCS representation as well as the domain-randomized and bidirectional aligned synthetic training pipeline. Comparing \textbf{Ours} against \textbf{Ours-no-tracking}, when motion tracking is utilized to provide visual feedback, the performance is dramatically boosted from 42.9\% to 100\%, indicating the benefit of continuously tracking the object state in high precision contact-rich tasks. With a tighter tolerance $0.5mm$, the performance gap between our approach and baseline methods becomes more significant, demonstrating superior precision and robustness. Finally, \textbf{Ours} remains feasible in solving the task with $0.1mm$ tolerance, but the success rate decreases to 28.6\%. We expect a further boosted performance by adding force feedback to the control policy \cite{gao2021kpam} or using advanced compliant control methods \cite{andrewbwrss}.









\subsection{Framework Analysis}\label{sec:analysis}

\noindent \textbf{Generalizability to scene configurations:} In addition to generalization across a category, it's also interesting to explore whether the skills learned from single visual demonstration generalizes to novel scene configurations not seen in the video. Fig. \ref{fig:analysis} (a) illustrates the initial gear poses relative to the receptacle in the demonstrated configuration along with the testing cases in the ``gear insertion'' experiments (Sec. \ref{sec:gear_task}). Fig. \ref{fig:analysis} (d) provides an example testing case for the battery assembly task where an obstacle impedes the direct transport towards the \textit{keypose}, which is unseen in the demonstration video (middle in first page's figure). In this case, a collision-free path is planned and executed. As observed, the approach generalizes to different scenes, including unstructured environments with novel obstacles. This is attributed to the separation of the long-range motion and last-inch manipulation, leveraging video parsing with object motion tracking (Sec. \ref{sec:tracking}). Thus, the method disentangles the object of interest from the background and represents the task-relevant trajectory independent of specific scenes.


\noindent \textbf{Robustness against external disturbance:} Fig. \ref{fig:analysis}(e) shows an example of gear insertion, which is successful despite external disturbances due to the human operator, who drags the object away from the gripper, causing a change of the gear's pose. With visual motion tracking constantly feeding the latest object  state to the controller (tracking visualizations shown in the bottom-left corners), the desired trajectory is closely followed, thus providing robustness to CatBC against disturbances.

\noindent \textbf{Endowing different manipulation strategies:} A key advantage of the proposed approach is the ability to conveniently endow the robot with various manipulation strategies, which is complicated to program otherwise. In the above $0.5mm$ tolerance ``gear insertion'' task (Sec. \ref{sec:gear_task}), a plain top-down insertion strategy is illustrated in the demonstration video (first page's figure). Then, a more reliable strategy is re-demonstrated using a different action sequence. Specifically, the gear's internal edge is first anchored against the shaft top, and the gear pivots around the anchored point. 
As shown in Fig. \ref{fig:analysis} (f), the demonstration encoding this strategy, named ``human pivot'', notably improves the success rate compared with the simple top-down strategy.


\noindent \textbf{Sensitivity to the demonstration format:} The demonstration video parsing formulation (Sec. \ref{sec:tracking}) not only disentangles the object of interest from the background, but also from the manipulator. Therefore, the framework smoothly works with other demonstration formats, such as kinesthetic teaching or tele-operation. Fig. \ref{fig:analysis} (b) includes evaluations in the ``gear insertion'' task with teleop-collected demonstration using the plain top-down  strategy (teleop naive). Compared to human arm demonstration, tele-operation uses the same robotic arm and thus does not suffer from the kinematics or compliance gap. Additionally, the robot arm motion is practically more stable than the human motion, providing a less noisy target trajectory. These lead to slightly improved performance.
However, for tasks involving complex long-horizon sequential actions, tele-operation might become cumbersome. In principal, the approach is not constrained to specific demonstration formats, and one can choose the format most suitable to the task.


\noindent \textbf{Running time:} The average running time of the last-inch manipulation is reported in Fig. \ref{fig:analysis} (c). The 6 DoF motion tracker provides visual feedback to CatBC in real time at the camera's frequency (10 Hz) and runs in a separate thread in parallel, adding nearly no delay other than the communication cost. The running time difference among the tasks is mainly due to the length of each target trajectory. In particular, for the ``battery assembly'' task, intricate long-horizon sequential actions are required and the CatBC process takes longer. For the long-range collision-aware motion, the running time primarily depends on the distance between the grasping pose and the \textit{keypose}, along with the complexity of the obstacles in the environment, thus omitted in the summary.



\section{DISCUSSIONS AND FUTURE WORK} 
\label{sec:conclusion}

This work presents a closed-loop category-level manipulation framework that uses visual feedback. The framework can be applied to novel objects given a single visual demonstration. Extensive experiments demonstrate its efficacy in a range of high-precision assembly tasks that require learning complex, long-horizon sequential policies. The approach provides robustness against uncertainty due to manipulation dynamics, and generalization across object instances and scenes. It also allows teaching a robot different manipulation strategies by solely providing a single demonstration, without the need for manual programming.


There are a few limitations that open up future work directions. First, the current framework utilizes vision as the single sensing modality. Integrating additional sensor modalities, such as force or tactile sensing, can further improve the accuracy of behavioral cloning for high precision manipulation tasks. In addition, only rigid objects are considered here. Many manipulation tasks involve articulated or deformable objects, such as cables, and suitable category-level representations are needed for such object categories. 

\section*{Acknowledgments}
Bowen Wen and Kostas Bekris were partially supported by the US NSF Grant IIS-1734492. The opinions expressed here are of the authors and do not reflect the views of the sponsor.  


\bibliographystyle{IEEEtran}
\bibliography{references}

\clearpage

\appendix

\subsection{Implementation details}

\begin{figure*}
\centering
\includegraphics[width = 0.9\textwidth]{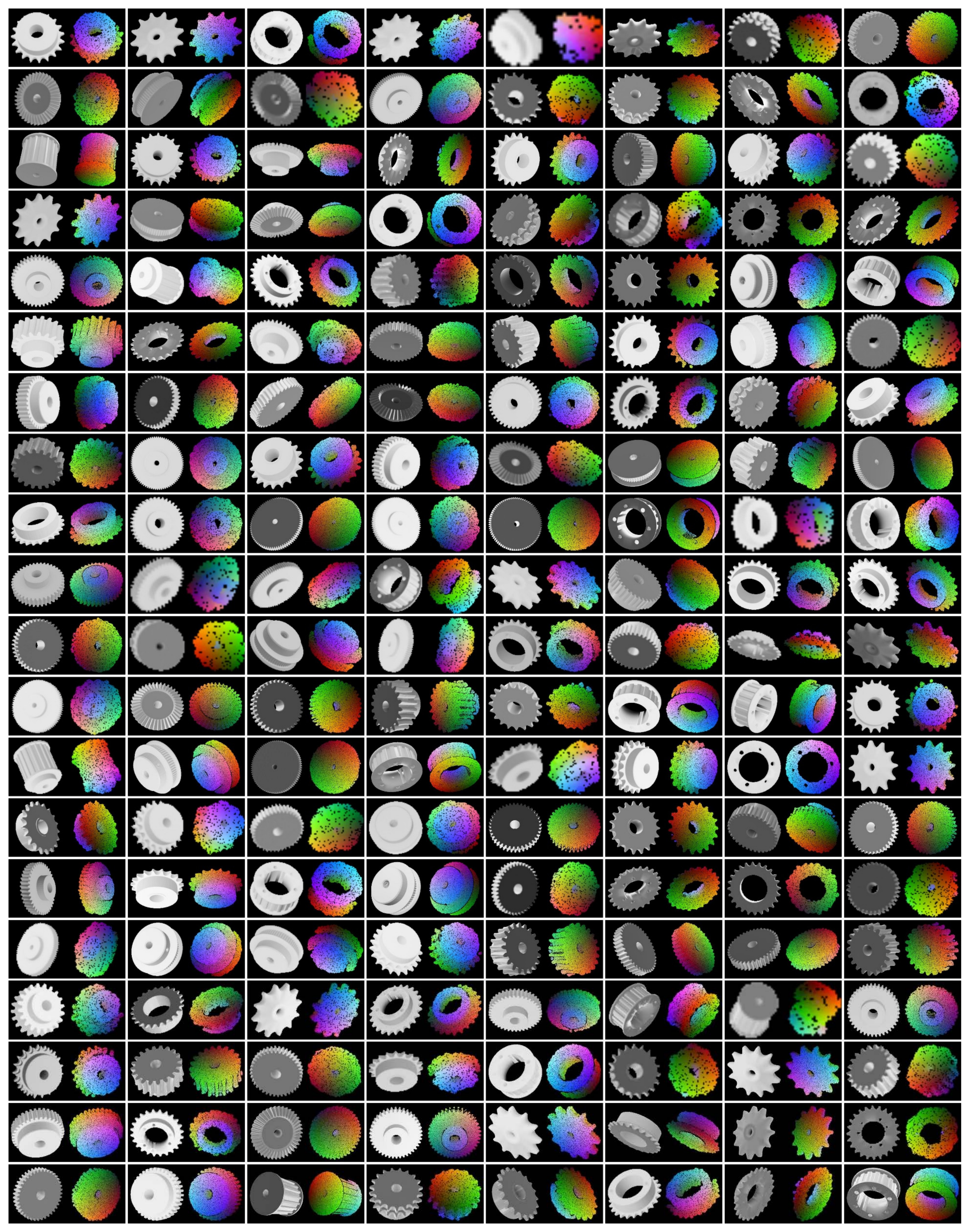}
\caption{Example synthetic training data in \textit{Gears} category to train the NUNOCS Net in our approach, where each pair consists of a color image and the ground-truth NUNOCS label corresponding to the corrupted noisy depth image. In total, 100000 data points have been generated in simulation, where each data point includes color and depth images, an instance segmentation mask, and the ground-truth NUNOCS label.}
\label{fig:ours_data_gear} 
\end{figure*}

\begin{figure*}
\centering
\includegraphics[width = 0.9\textwidth]{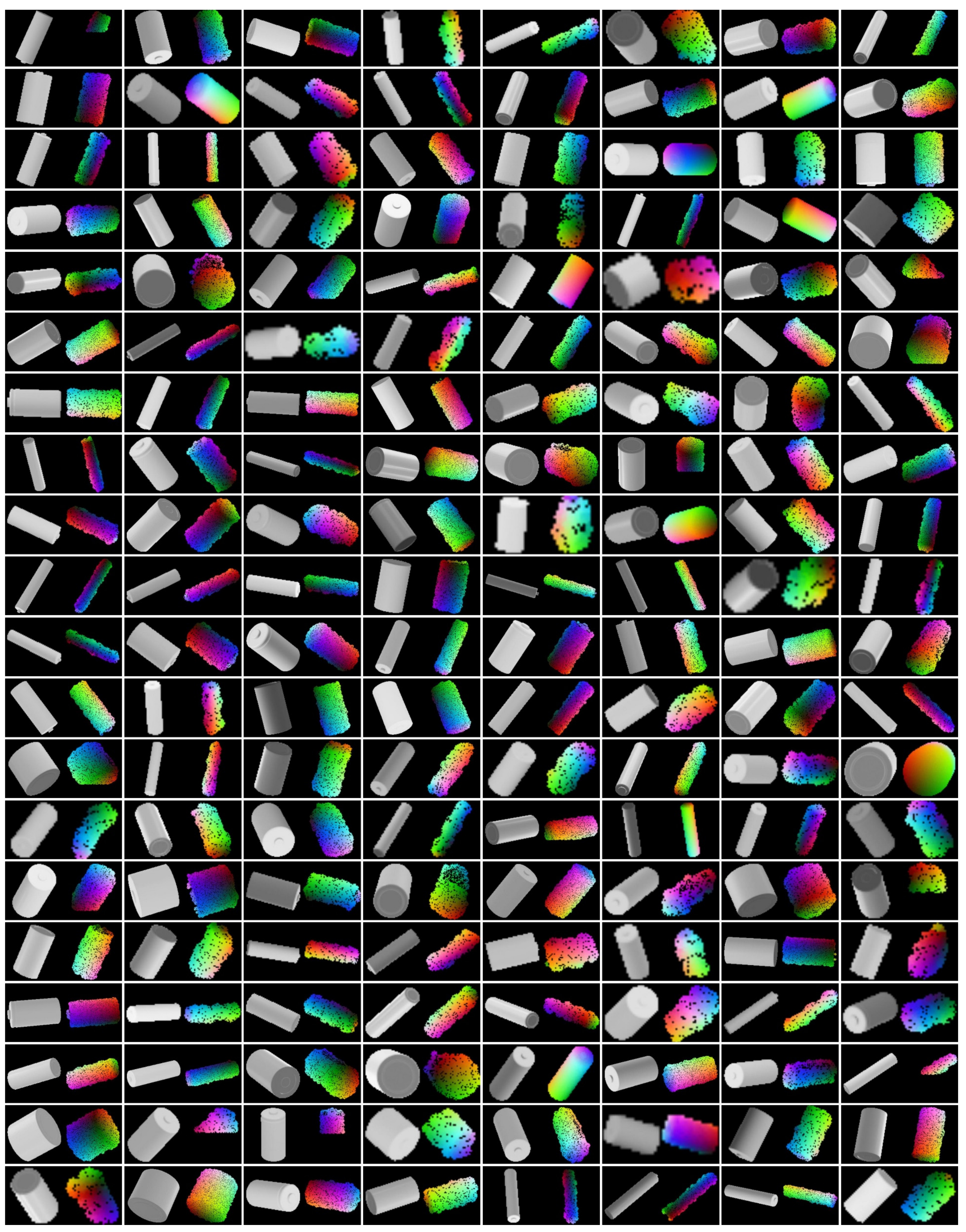}
\caption{Example synthetic training data in \textit{Battery} category to train the NUNOCS Net in our approach, where each pair consists of a color image and the ground-truth NUNOCS label corresponding to the corrupted noisy depth image. In total, 100000 data points have been generated in simulation, where each data point includes color and depth images, an instance segmentation mask, and the ground-truth NUNOCS label.}
\label{fig:ours_data_battery} 
\end{figure*}

The NUNOCS learning process is solely conducted over synthetic data. Example training data is exhibited in Fig. \ref{fig:ours_data_gear} and Fig. \ref{fig:ours_data_battery}. The data generation pipeline is developed with Blender \cite{blender2018}. For each data point generation, an object is randomly selected from the training set and dropped onto a table surface. After the object stabilizes, the partially scanned point cloud is computed from the rendered 2D depth image. Additionally, the object’s ground-truth pose is retrieved along with its CAD model to compute the training labels $\overline{\mc P}_\mathbb{C}$ and $\overline{s}$. The object type, initial pose relative to the camera, table's height, gravity, friction and bouncing parameters are all randomized in each scene. Additionally, random non-uniform 3D scaling is applied to the object model to attain a novel object with different dimensions, which allows to create greatly enriched kinds of object instances beyond the ones provided in the training set. Each data point includes the scanned depth image, instance segmentation mask and the ground-truth labels of NUNOCS. The NUNOCS Net is implemented in Pytorch \cite{paszke2019pytorch} and is trained with Adam optimizer for 100 epochs with a batch size of 50. Learning rate starts from 0.01 and is scaled by 0.1 at epochs 50 and 80. Depth-missing corruption is applied to the 2D depth image at a missing percentage between 0 to 0.4 before being converted into the point cloud. Additional data augmentations include random translation and rotation applied to the point cloud. 

For the trajectory extracted from the single demonstration video, timestamps are discretized such that the interval distance between any neighboring sub-goals are at least $2mm$ or $2\degree$. This discretizes the continuous trajectory while ensures to timely correct from deviations when performing the CatBC (Sec. \ref{sec:catbc}). One thing noteworthy is symmetry handling when following the target trajectory during online CatBC. For both the categories of \textit{Gears} and \textit{Batteries}, the equivalent symmetric transformations are the rotations around the Z axis from 0 to $360\degree$ discretized by $5\degree$. In such cases, for each of the intermediate goals, each equivalent transformed pose, sorted by their distances to the current object pose in ascending order, is sequentially checked for a collision-free IK solution. The first feasible equivalent pose is then selected as the next sub-goal pose.

\subsection{Limitations and failure modes} 
\begin{figure*}[t]
\centering
\includegraphics[width = 0.9\textwidth]{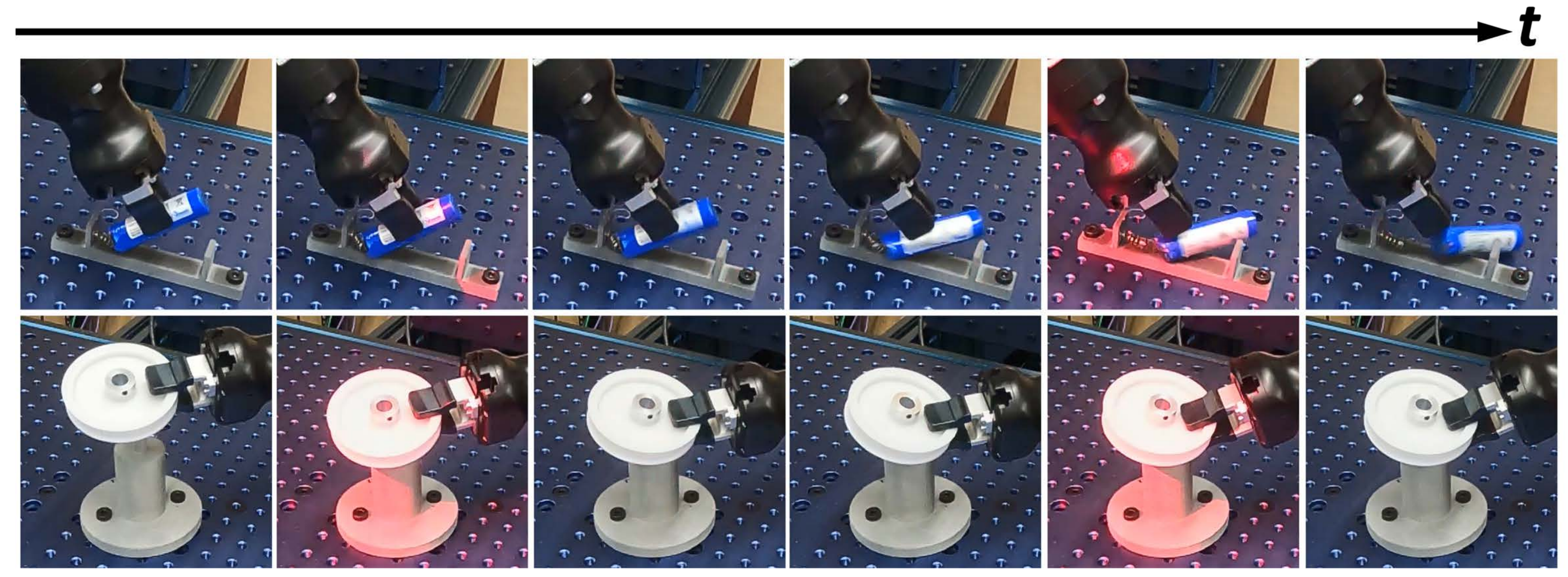}
\caption{Example failure modes. \textbf{Top:} In the ``battery assembly'' task, as the battery squeezes the spring, the elastic force gradually increases and eventually pushes the battery out of the gripper. In this case, it would be beneficial to add tactile sensing together with the visual tracker in the feedback loop to predict slippery early.  \textbf{Bottom:} In the ``gear insertion'' task, when using “anchor-and-pivot” strategy, rich contact leads to significant change of the gear's in hand orientation, causing no collision-free IK solutions found to continue. In this scenario, the object is still accurately tracked, so regrasping or adjusting the object pose via in-hand manipulation with a dexterous hand can recover from this failure mode.}
\label{fig:failure} 
\end{figure*}

Example failure modes are presented in Fig. \ref{fig:failure}. In the ``battery assembly'' task (top), as the battery squeezes the spring, the elastic force gradually increases and eventually pushes the battery out of the gripper. In this case, it would be beneficial to add tactile sensing together with the visual tracker in the feedback loop to predict slippery early. In the ``gear insertion'' task (bottom), when using “anchor-and-pivot” strategy, rich contact leads to significant change of the gear's in hand orientation, causing no collision-free IK solutions found to continue. In this scenario, the object is still accurately tracked, so regrasping or adjusting the object pose via in-hand manipulation with a dexterous hand can recover from this failure mode.

\subsection{Baseline methods}
\textbf{DON} - This is based on the open-sourced implementation\footnote{\url{https://github.com/RobotLocomotion/pytorch-dense-correspondence}}  of \cite{florence2018dense}. The same training data collection pipeline in \cite{florence2018dense} is adopted, where dense correspondence across object instances is learned through contrastive learning, without requiring annotated labels. Some example training data collected on our task objects are displayed in Appendix.
We closely follow the semantic grasping learning workflow in \cite{florence2018dense} and specify semantic keypoints that are required to define the manipulation task target. For fair comparison, the semantic keypoints are specified on images corresponding to the same $\mc{O_D}$ in our used demonstration video. The keypoints are then transferred to novel object instances during testing based on the learned cross-image dense correspondence.

\textbf{DON BC} - This is a modified version of \textbf{DON} \cite{florence2018dense} by replacing the manual goal specification with the single visual demonstration, similar to ours. The same video parsing
and CatBC formulation proposed in our approach are integrated, except that the cross-image dense correspondence learned from real world data is inherited from \textbf{DON} instead of our category-level representation for the CacBC process. When directly using the predicted dense correspondence for parsing the in-hand object motion in the demonstration video, where significant occlusions involve, the trajectory quality is poor. Therefore, the same extracted trajectory by our approach is utilized. During testing, the predicted dense correspondence initializes the object's state and object motion is thereafter tracked for CatBC by treating them as additional virtual links in the kinematic tree, same as \cite{manuelli2019kpam}.

\textbf{KPAM} - This is based on the open-sourced implementation\footnote{\url{https://github.com/weigao95/kPAM}} of \cite{manuelli2019kpam}. The training data collection is similar to that of \textbf{DON} \cite{florence2018dense}. Following the original work \cite{manuelli2019kpam}, human demonstration is provided via manual goal specification during the testing stage for trajectory optimization.

\textbf{KPAM BC} - This is a modified version of \textbf{KPAM} \cite{manuelli2019kpam} augmented with behavior cloning by replacing the manual goal specification with the single visual demonstration, similar to ours. The same video parsing and CatBC formulation proposed in our approach are integrated similar to the treatment to \textbf{DON BC}, except that during testing the cross-instance correspondence is established using the \textbf{KPAM} predicted semantic keypoints for the CatBC process. Similar to the reason in \textbf{DON BC}, the same extracted trajectory by our approach is utilized. Strictly following \textbf{KPAM}, the keypoints are tracked by treating them as additional virtual links in the kinematic tree \cite{manuelli2019kpam}.

\textbf{KPAM 2.0} - This is based on the related work \cite{gao2021kpam} that extends \textbf{KPAM} \cite{manuelli2019kpam} by augmenting semantic keypoints with orientation information for improved expressiveness. 
The force sensing is disabled to make vision as the primary sensing modality, sharing the same setup as other evaluated approaches.

\textbf{KPAM 2.0 BC} - This is a modified version of \textbf{KPAM 2.0} \cite{gao2021kpam} by replacing the manual goal specification with the single visual demonstration, similar to ours. The same video parsing and CatBC formulation proposed in our approach are integrated similar to the treatment to \textbf{DON BC}, except that during testing the cross-instance correspondence is established from \textbf{KPAM 2.0}'s predicted semantic keypoints for the CatBC process. Similar to the reason in \textbf{DON BC}, the same extracted trajectory by our approach is utilized. The keypoints are tracked kinematically in the same fashion as in \textbf{KPAM BC}.

Besides the above category-level manipulation approaches, another concurrent work \cite{simeonov2021neural} requires 4 RGBD cameras mounted at each corner on a tabletop, which allows to fuse the point cloud to reconstruct the 3D shape of the novel objects. This is different from the single camera setting considered in all the evaluated approaches and thus not included for comparison.

\subsection{Training data collection for baseline methods} 
\label{sec:training_data_baseline}

\begin{figure*}
\centering
\includegraphics[width = 0.9\textwidth]{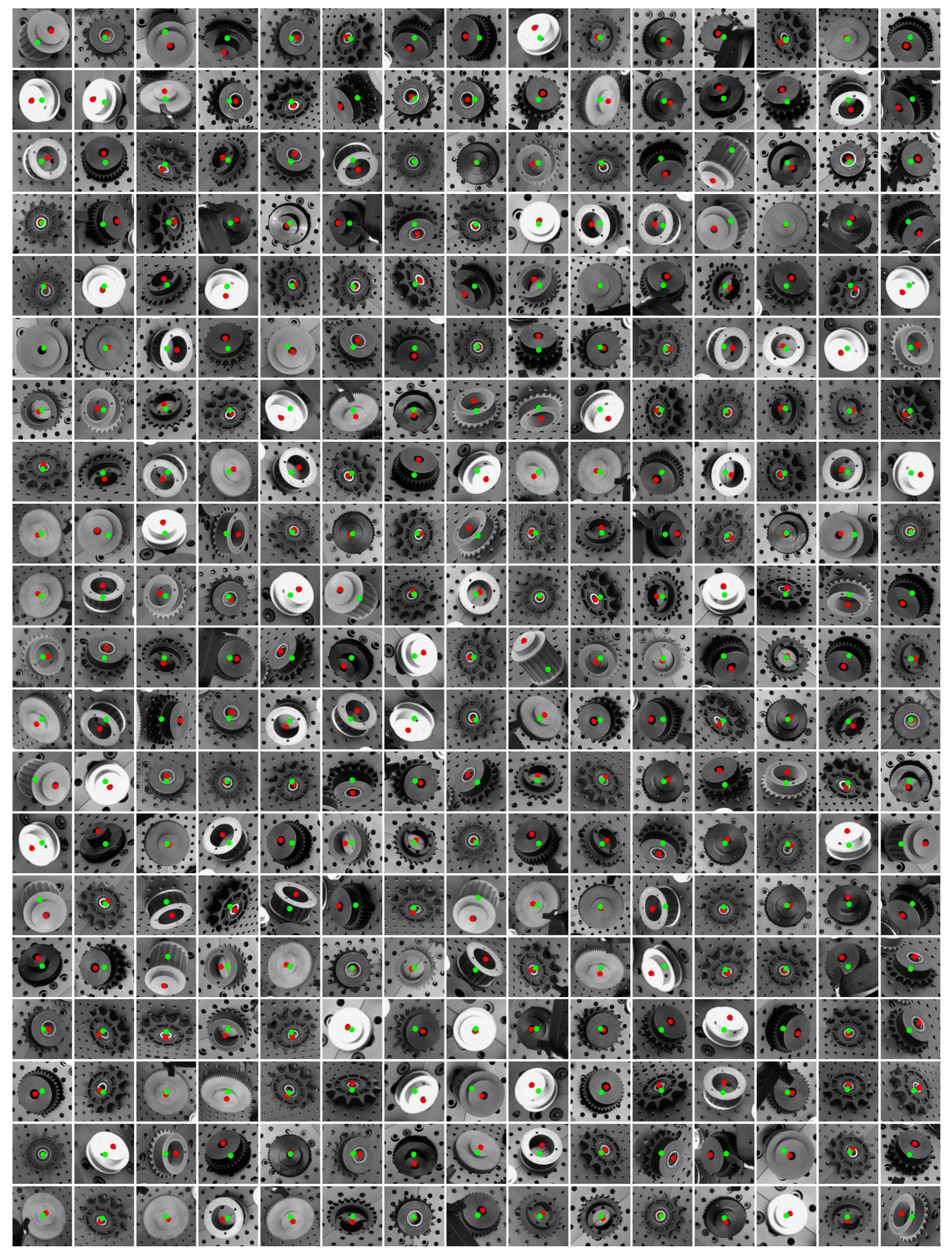}
\caption{Example real-world training data in \textit{Gears} category collected in this work to train KPAM \cite{manuelli2019kpam} for comparison. In total, 98340 data points have been generated, where each data point includes color and depth images, an object bounding box, an instance segmentation mask, and ground-truth semantic keypoints annotated in red and green.}
\label{fig:kpam_data_gear} 
\end{figure*}

\begin{figure*}
\centering
\includegraphics[width = 0.9\textwidth]{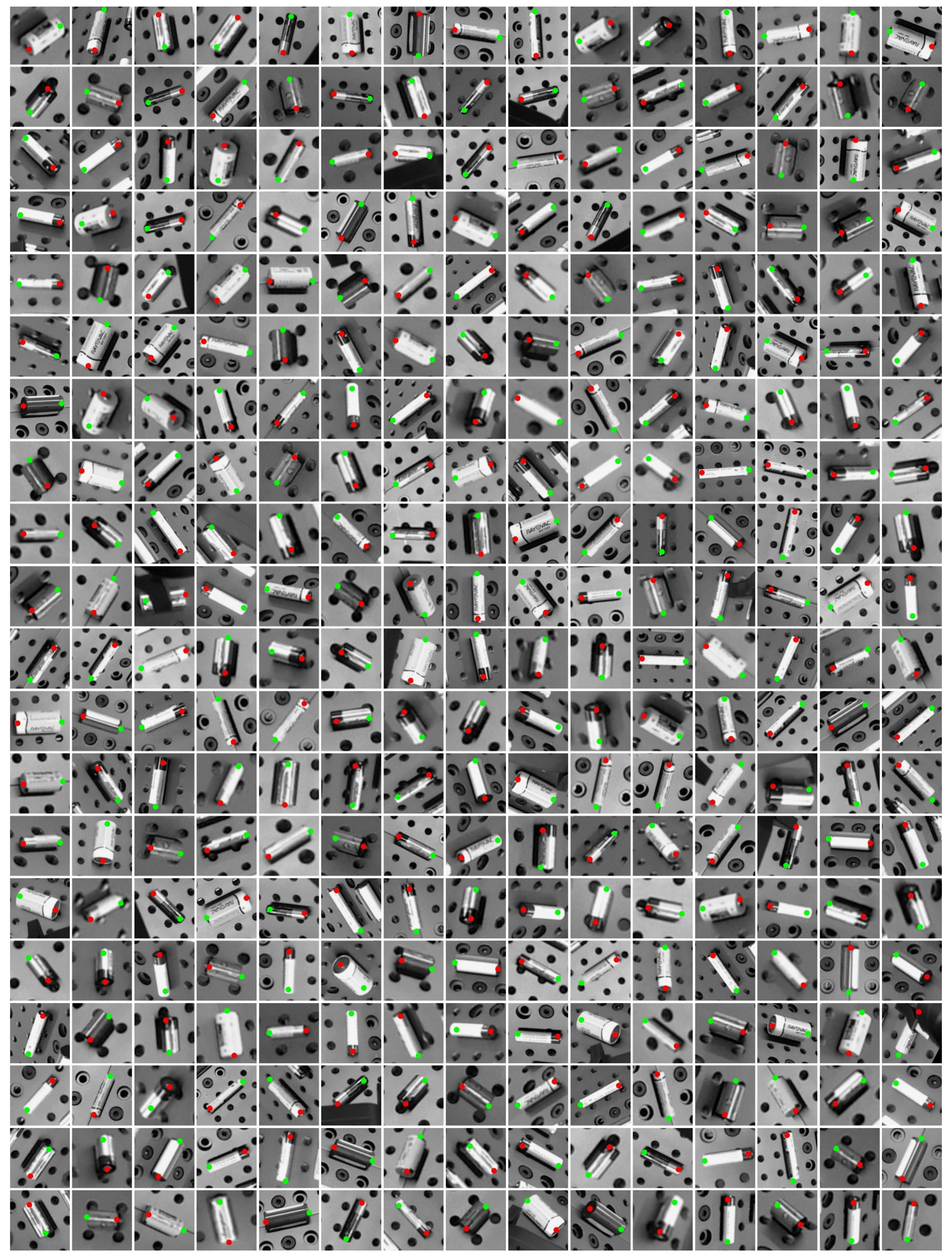}
\caption{Example real-world training data in \textit{Batteries} category collected in this work to train KPAM \cite{manuelli2019kpam} for comparison. In total, 90000 data points have been generated, where each data point includes color and depth images, an object bounding box, an instance segmentation mask, and ground-truth semantic keypoints annotated in red and green.}
\label{fig:kpam_data_battery} 
\end{figure*}

\begin{figure*}
\centering
\includegraphics[width = 0.9\textwidth]{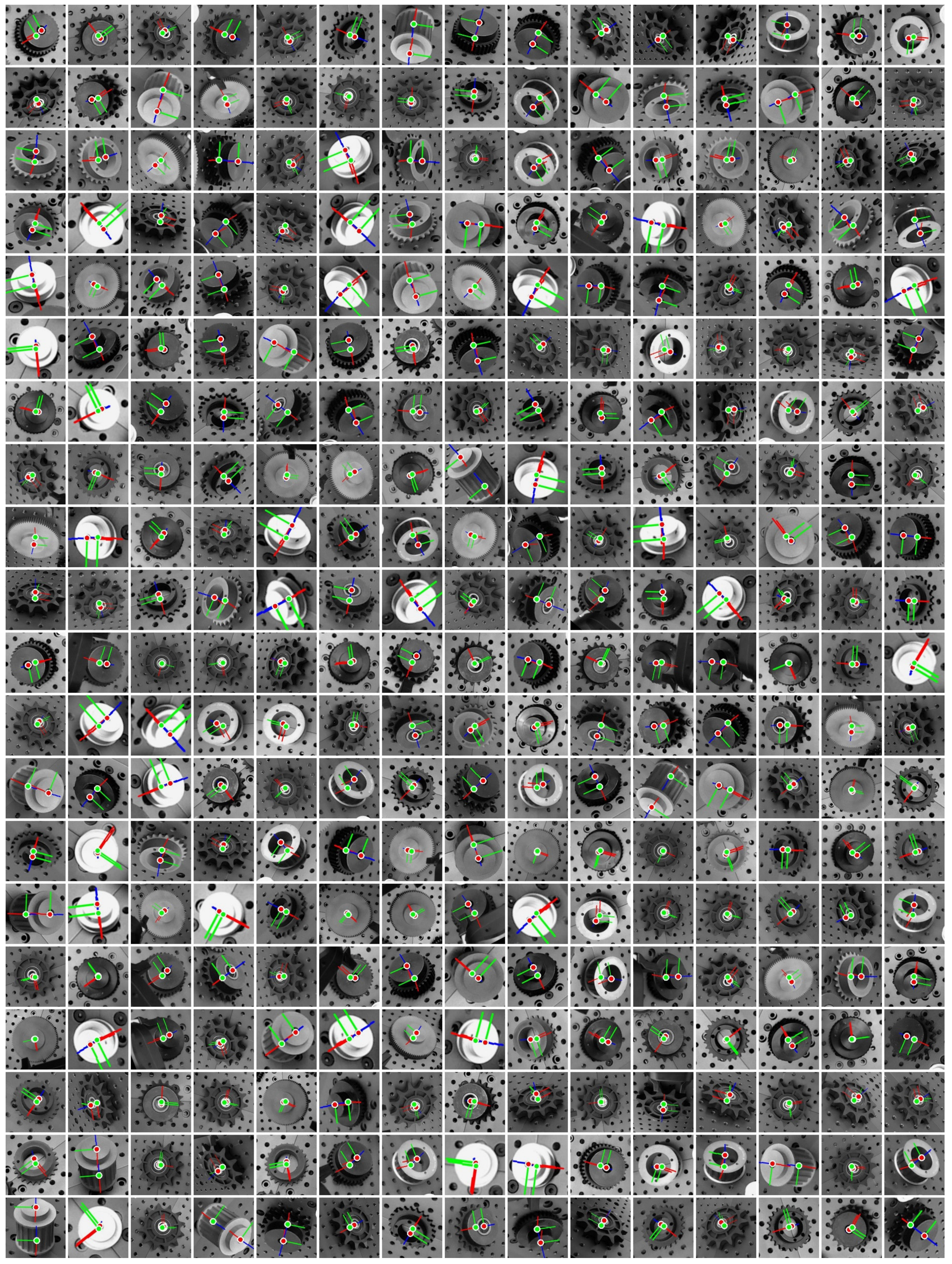}
\caption{Example real-world training data in \textit{Gears} category collected in this work to train KPAM 2.0 \cite{gao2021kpam} for comparison. In total, 90000 data points have been generated, where each data point includes color and depth images, an object bounding box, an instance segmentation mask, ground-truth semantic keypoints annotated in red and green, and their augmented orientations.}
\label{fig:kpam2_data_gear} 
\end{figure*}

\begin{figure*}
\centering
\includegraphics[width = 0.9\textwidth]{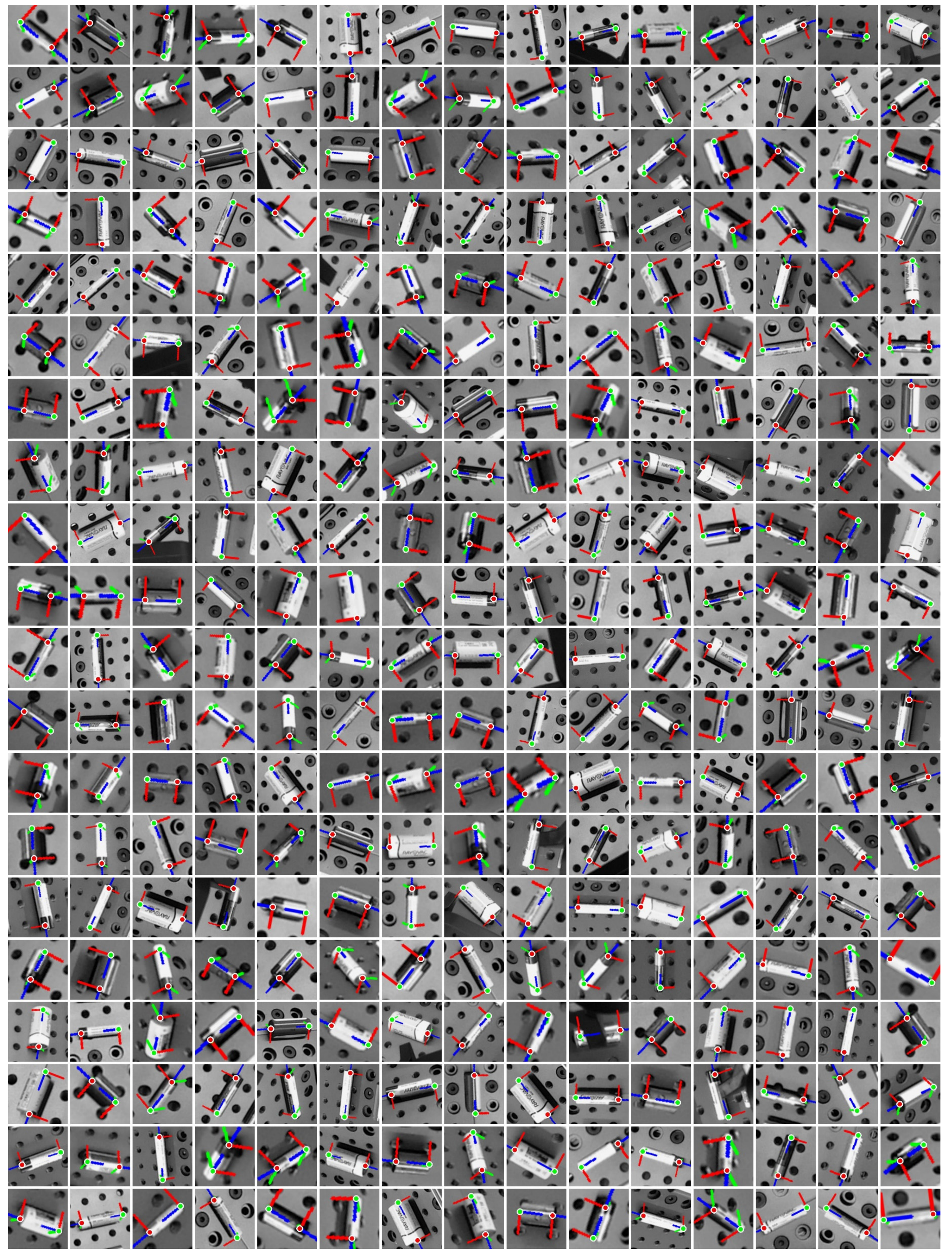}
\caption{Example real-world training data in \textit{Batteries} category collected in this work to train KPAM 2.0 \cite{gao2021kpam} for comparison. In total, 90000 data points have been generated, where each data point includes color and depth images, an object bounding box, an instance segmentation mask, ground-truth semantic keypoints annotated in red and green, and their augmented orientations.}
\label{fig:kpam2_data_battery} 
\end{figure*}

\begin{figure*}
\centering
\includegraphics[width = 0.9\textwidth]{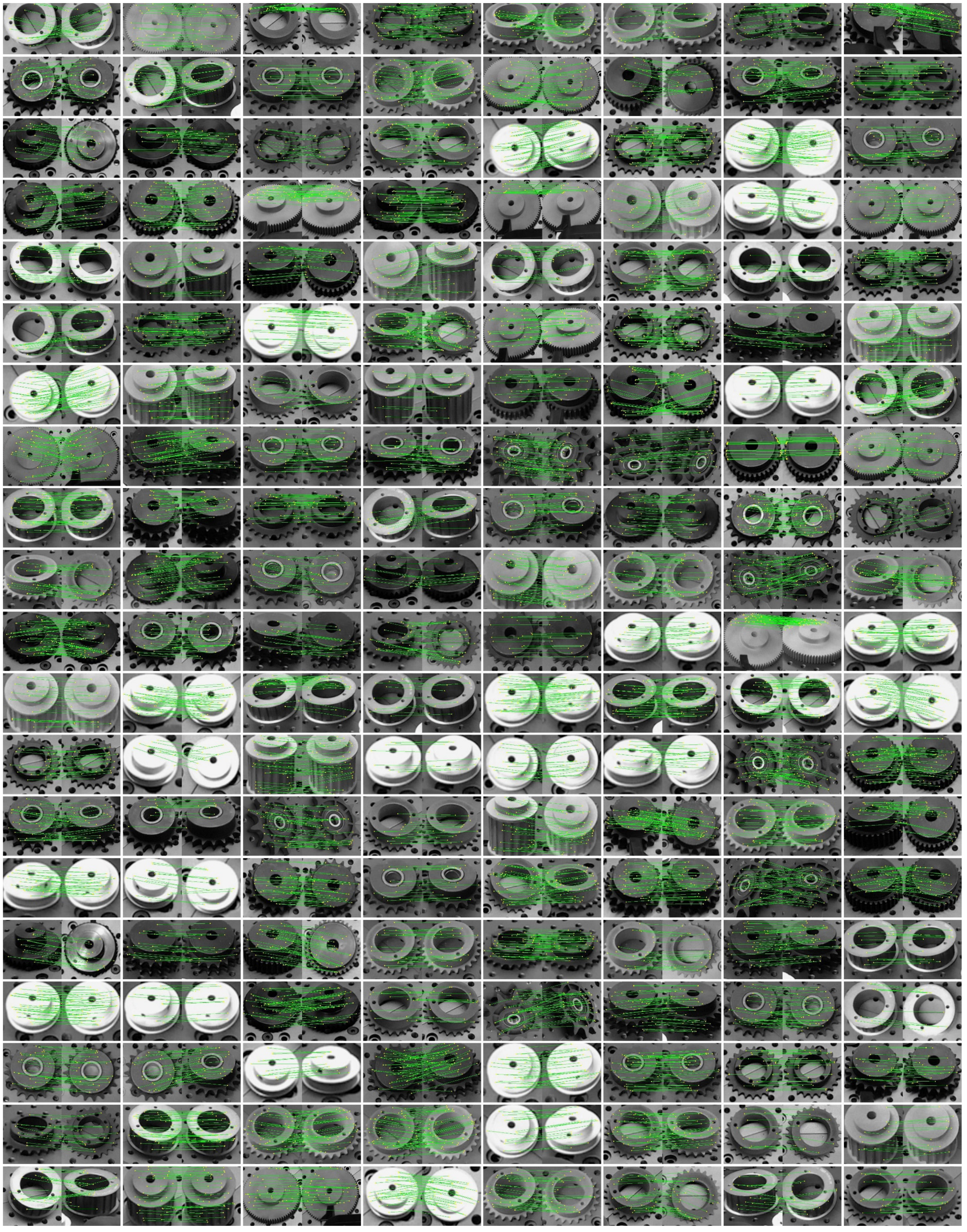}
\caption{Example real-world training data illustrating sampled inlier correspondences in \textit{Gears} category collected in this work to train the comparison approach DON \cite{florence2018dense}. During training, DON randomly samples inlier and outlier correspondences for contrastive learning.}
\label{fig:don_data_gear} 
\end{figure*}

\begin{figure*}
\centering
\includegraphics[width = 0.9\textwidth]{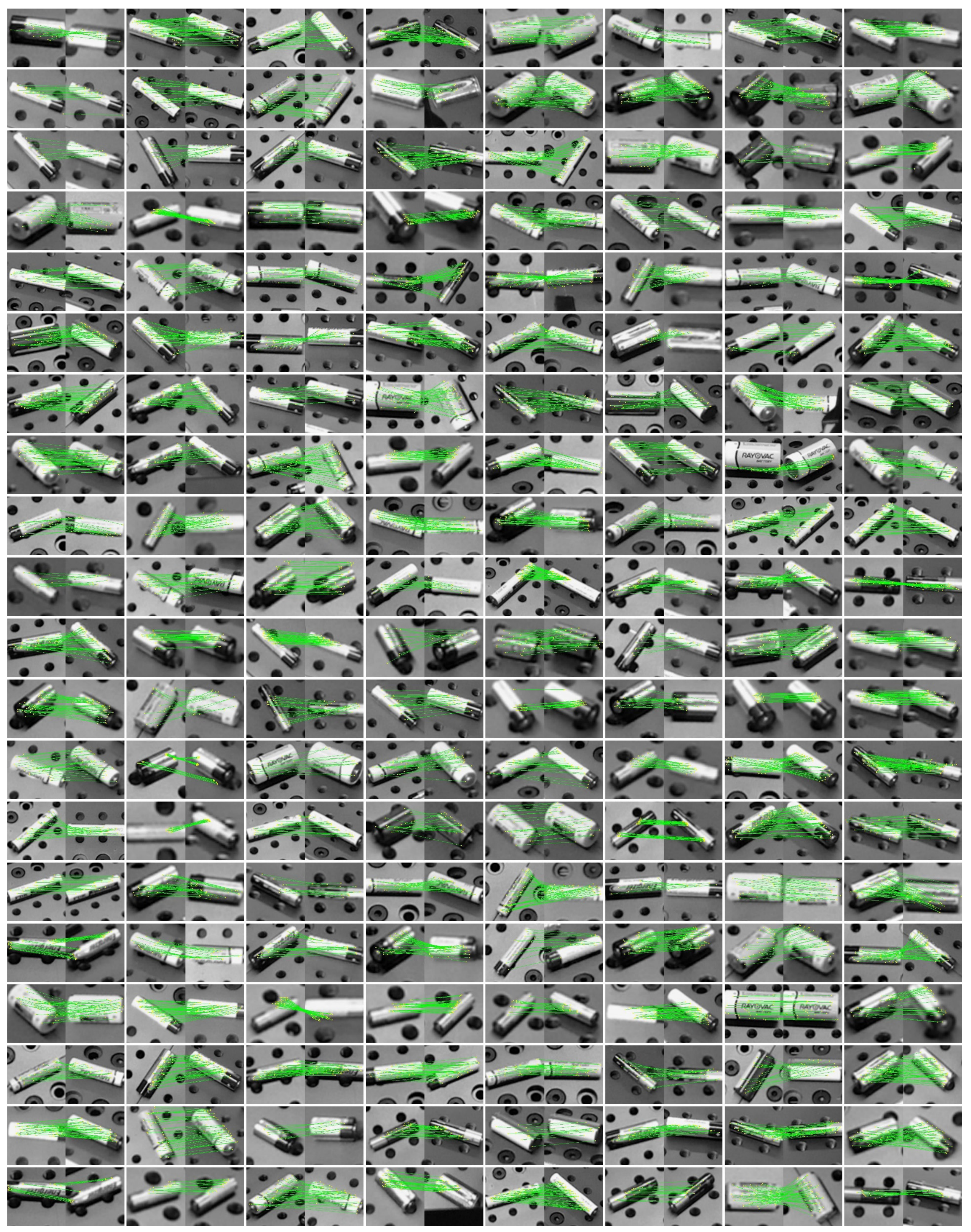}
\caption{Example real-world training data illustrating sampled inlier correspondences in \textit{Batteries} category collected in this work to train the comparison approach DON \cite{florence2018dense}. During training, DON randomly samples inlier and outlier correspondences for contrastive learning.}
\label{fig:don_data_battery} 
\end{figure*}

Data collection process described in \cite{florence2018dense,manuelli2019kpam,gao2021kpam} was closely followed. For \textbf{KPAM} \cite{manuelli2019kpam}, some example training data with keypoint annotations are shown in Fig. \ref{fig:kpam_data_gear} and \ref{fig:kpam_data_battery}. For \textbf{KPAM 2.0} \cite{gao2021kpam}, some example training data with keypoint annotations augmented with orientations are shown in Fig. \ref{fig:kpam2_data_gear} and \ref{fig:kpam2_data_battery}. For \textbf{DON} \cite{florence2018dense}, some example training data with sampled inlier correspondences obtained by re-projecting points based on their relative transformations is shown in Fig. \ref{fig:don_data_gear} and \ref{fig:don_data_battery}.

\end{document}